\documentclass[10pt,twocolumn,journal final]{IEEEtran}
\usepackage{graphicx}
\usepackage{algorithm}
\usepackage{algorithmic}
\usepackage{subfig}
\usepackage[cmex10]{amsmath}
\usepackage{ amssymb }
\usepackage{amsthm}
\usepackage{overpic}
\newtheorem{proposition}{Proposition}
\newtheorem{proposition2}{Proposition}

\begin{document}
\abovedisplayskip=6pt
\belowdisplayskip=6pt

\title{Efficient MRF Energy Propagation for Video Segmentation via Bilateral Filters}
\author{Ozan~\c{S}ener, Kemal Ugur and A. Ayd\i n Alatan,~\IEEEmembership{Senior Member,~IEEE}}


\maketitle

\begin{abstract}
Segmentation of an object from a video is a challenging task in multimedia applications. Depending on the application, automatic or interactive methods are desired; however, regardless of the application type, efficient computation of video object segmentation is crucial for time-critical applications; specifically, mobile and interactive applications require near real-time efficiencies. In this paper, we address the problem of video segmentation from the perspective of efficiency. We initially redefine the problem of video object segmentation as the propagation of MRF energies along the temporal domain. For this purpose, a novel and efficient method is proposed to propagate MRF energies throughout the frames via bilateral filters without using any global texture, color or shape model. Recently presented bi-exponential filter is utilized for efficiency, whereas a novel technique is also developed to dynamically solve graph-cuts for varying, non-lattice graphs in general linear filtering scenario. These improvements are experimented for both automatic and interactive video segmentation scenarios. Moreover, in addition to the efficiency, segmentation quality is also tested both quantitatively and qualitatively. Indeed, for some challenging examples, significant time efficiency is observed without loss of segmentation quality.
\end{abstract}


\IEEEpeerreviewmaketitle
\section{Introduction}
Segmenting an object of interest from an image or a video is a crucial task in many multimedia applications. Therefore, this problem has received significant interest for a long time. Moreover, both unsupervised and semi-supervised version of the problem has been attacked by many researchers. 

Although there exist many techniques for automatic image segmentation \cite{NCut,FH,IsoPerimetric}, the problem is considered to be ill-posed due to the ambiguous definition of an "object" and lack of prior information for a successful object segmentation. On the other hand; high quality, interactive image segmentation is possible due to the recent improvements. Generally, by small supervision (bounding box of the object \cite{grabcut}, approximate boundary of the object \cite{ls}, a few scribbles on foreground and/or background \cite{boykovSegmentation,paintselection}), precise object boundaries or matte values can be obtained. Interactive image segmentation problem is generally converted to a two-label energy minimization problem and object boundaries are obtained via energy minimization \cite{boykovMethod} while color and texture cues of the interacted pixels are used to define this energy function \cite{grabcut}. Coherency of the segmentation result is also enforced via smoothness penalty term \cite{boykovSegmentation}. The most common energy function used in interactive image segmentation scenario is a Markov Random Field (MRF) energy \cite{boykovSegmentation,grabcut,paintselection,ls}. MRF energies are minimized via min-cut/max-flow method efficiently \cite{boykovMethod}. By the introduction of the dynamic methods \cite{mmpd,dynGCut} and hierarchical graph-cut \cite{hGCut}, even near real time execution is possible under some constraints. Moreover, mobile applications are also emerging as a result of these improvements \cite{mmpd}.

Adapting such image segmentation methods directly to the video segmentation problem is not feasible due to the additional temporal dimension, as well as the necessity to use motion information for a high quality segmentation. First of all, by the help of the motion information, it is possible to redefine the segmentation problem and solve it by an fully automatic method. However, ambiguity of the object definition still exists; object of interest might not be moving in the sequence. In order to solve this drawback, a definition of the object is revisited by the help of saliency and motion information \cite{alatan94,whatObject,bplr}. Saliency-based definition of the object has already led to successful automatic video segmentation applications \cite{keysegments}. Although the results of this algorithm looks quite promising, both motion estimation and saliency measure are computationally expensive processes. Therefore, it is almost impossible to apply them to any interactive or mobile multimedia application.

Bottom-up processing of a video is another approach to segment videos. Bottom-up methods start with redundant over-segmentation of a video, and apply spatio-temporal clustering to obtain segmentations \cite{graphVideo,MHVS}. In \cite{graphVideo}, nodes of the 3D graph is obtained via spatio-temporal over-segmentation; and, edges are added via computed optical flow. Then, clustering is performed based on the desired level of hierarchy. Using supervision for the hierarchical level selection makes this method suitable for many interactive multimedia scenarios; however, computationally expensive extraction of motion information make it impossible to apply such scenarios. In another bottom-up method \cite{MHVS}, 2D over-segmentation of each frame is obtained. And, over-segment flows are computed via inference on a Markov chain to cluster over-segments. However, inference on Markov chains is also a computationally expensive procedure; therefore, this method is also not applicable to interactive and mobile multimedia applications.

On the other hand, interactive video segmentation methods generally utilizes interactions in some keyframes \cite{roto,geos,cutPaste}. After interaction, an energy function exploiting texture, color and interaction is defined and minimized \cite{cutPaste,intCutOut}. However, these approaches generate high quality segmentation results for a few frames only and require user interactions in many other frames. The main reason for this performance drawback is the lack of motion information, which is computationally quite expensive to be used. Therefore, it is desired to avoid motion information for the applications demanding efficiency. However, not only motion information, but also spatial relations are discarded in these methods due to the global models \cite{geos,cutPaste}. It should be noted that these methods are the extensions of interactive image segmentation algorithms and most of the interactive image segmentation algorithms use global models, such as GMM \cite{grabcut} or Kernel Density Estimation \cite{geo2,kde}. Moreover, global color and texture models discard the spatial information. A naive solution to this problem is obtained by using a spatial distance aware metric, such as \emph{geodesic distance}. In \cite{geos}, spatio-temporal geodesic distance of each pixel to interaction points are computed and used in a graph-cut scenario. Although, this method significantly increases the number of frames that can be segmented without extra user interaction, geodesic distances are not robust against color profile changes or high amount of motion. Therefore, user interaction might be required in many frames which decrease the interaction quality significantly. On the other hand, in \cite{roto}, energy function is defined over local overlapping windows. Therefore, locality of the energy function is significantly emphasized. These local windows are propagated by using the computed optical flow information. As a motion information-aware method, the main drawback of this method is its time complexity \cite{roto}.

In summary, fully automatic video segmentation methods require computationally expensive computation of motion information for a successful operation. On the other hand, interactive methods exploit computationally efficient algorithms producing low quality video segmentation results and these methods try to overcome possible segmentation quality drawback via extensive and redundant amount of user interaction. However, extensive amount of user interaction reduces the quality of the user experience. When an interactive multimedia application is considered, efficiency, segmentation quality and quality of user interaction are all equally crucial. Therefore, the interaction should be at the minimum level, whereas the algorithm should be efficient and the results should be accurate. Indeed, these requirements pose a serious dilemma for video segmentation problems in multimedia applications. 

In this paper, in order to solve this dilemma, we redefine the video object segmentation problem as an estimation of the foreground probability density function of each frame in terms of probability density functions of previous frames. By the help of Markovian property assumption, we can define this problem as estimating the MRF energy of each frame in terms of MRF energy for the previous frame. In our framework, we assume that MRF-based energy function for the first frame is already known. In practice, this unknown energy function can be obtained via user interaction or estimating the optical flow between the first two frames. We propose an efficient method to propagate energy distributions throughout the frames via bilateral filters. The weights of these filters are selected by using texture similarities and spatio-temporal relations. Moreover, an increase in the segmentation quality is expected by the help of simultaneous usage of spatial and texture cues. In order to increase the efficiency of the method further, we also propose a dynamic algorithm for the solution of the min-cut/max-flow problem in bilateral filtering scenario. The proposed method does not use any global shape or color model, it also does not use motion information. We compensate lack of motion information via extensive usage of spatial information during the calculation of bilateral filter weights. Indeed, for the videos having low amount of motion, the proposed method results in high quality segmentation even for non-rigid object motion. The proposed method is also general enough to convert any MRF-based interactive image segmentation algorithm to an interactive video segmentation method. Moreover, it can also be used to speed-up any automatic video segmentation tool.

Our major contributions in this paper can be summarized as: 1) Utilization of bi-exponential filters to estimate the MRF energy of each frame from a consecutive previous frame that yields efficient and high quality video object segmentation; 2) Efficient and dynamic method to solve min-cut/max-flow in linear filtering scenario via linear recycling of residual flows.

The proposed algorithm is explained in three main parts: In Section II, a general method to estimate the MRF energy of a frame by using MRF energy of previous frame is proposed via spatio-temporal bilateral filtering. In Section III, an efficient approximation of geodesic bilateral filter is introduced to increase the computational efficiency. Finally, in Section IV, a dynamic method to efficiently solve graph-cuts in linear filtering scenarios is proposed.

\section{Estimation of MRF Energies}
The proposed framework solves video segmentation problem for each frame separately in order to reduce the dimension since solving 2D graph-cuts is more efficient than solving 3D spatio-temporal graph-cuts. Hence, we can define video segmentation problem as estimating the MRF energy of the current frame by using the MRF energy of the previous frame. In other words, the proposed estimation methodology propagates the energy functions throughout the frames. Before starting to explain the details of this propagation method, we should summarize the usage scenario for this propagation. The proposed method can either be used as an interactive video segmentation tool or speed-up tool for an automatic video segmentation. Any MRF energy minimization-based interactive image segmentation method \cite{grabcut,mmpd,boykovSegmentation} can be used to generate MRF energy of the initial frame. Then, the obtained energy function can be propagated via the proposed algorithm. Hence, the overall algorithm can be used as an interactive segmentation tool. For an automatic video segmentation case, any optical flow based method exploiting MRF energies can be used to segment the initial frame \cite{graphVideo}; then, the resultant energy can be propagated via the proposed method to speed-up the algorithm for the rest of the video.

In order to explain the method which estimates MRF energies in terms of the MRF energy of the previous frame, we first need to state the explicit form of the energy function. As explained in \cite{what_ef}, there exist a particular class of MRF energies which can be optimized efficiently via min-cut/max-flow method; namely submodular energy functions \cite{what-ef}. We are specifically interested in this class of energy functions. Generally, these energies composed of two main parts; the first part is the unary term which represents the consistency of the segmentation labels against some predefined models. The other term is the binary one and represents the coherency of the segmentation result. MRF energy over a graph, $G(\mathcal{V}, \mathcal{E})$, can be represented as
\[
E(\alpha)=\sum_{v_i \in \mathcal{V}} U(\alpha_i,z_i) + \sum_{v_i \in \mathcal{V}} \sum_{v_j \in N(v_i)} V(z_i,z_j) \phi[\alpha_i \neq \alpha_j]
\]
Although this formulation is usually applied to image data whose each pixel is a node, in order to increase the efficiency, over-segmenting an image is also a common technique \cite{MHVS,mmpd,graphVideo}. Hence, we will refer to nodes as regions in the image; these regions might correspond to pixels or over-segments depending on the application. In this representation, $\alpha_i$ represents the label of the region $i$, $z_i$ represents the color, shape and/or motion information related to region $i$ and $N(v_i)$ represents the neighbours of region $i$. $\phi[x]$ is an indicator function, yields one, if $x$ is true and 0 otherwise. $U(\alpha_i,z_i)$ is the unary energy corresponding to feature vector $z_i$ and label $\alpha_i$. Moreover, $V(z_i,z_j)$ term represents the penalty associated for giving different labels to two neighbour regions, where $\alpha$ represents the concatenation of labels of each region in the image. As a clarification to the notation, $\mathcal{V}$ corresponds to set of nodes (set of $v_i$) and $V(v_i,v_j)$ corresponds to the binary energy terms associated with node $i$ and $j$.

When we consider the limited amount of camera, as well as object motion, we can assume that there exist a matching region (i.e. a region with a similar appearance) in the previous frame for each region in the current frame, except for the occlusion regions. Therefore, one can argue that MRF energy of the current frame can be approximately represented in terms of the MRF energy of the regions in the previous frame. 

In order to proceed further, one should assume that there exist a distance metric between each region in the current frame and the previous frame. This distance metric should correspond to the spatio-temporal distance and texture similarity, simultaneously. Among alternatives, geodesic distance is a suitable metric to be used in such setting. Geodesic distance can be considered as the minimum cost among the paths connecting two specified regions. In the proposed framework, this cost can be defined as the weighted sum of the color differences along the spatio-temporal path and length of the path. A detailed analysis of the geodesic distances and their usage scenarios in image/video segmentation problem can be found in \cite{geos}. In the proposed method, an approximation to the geodesic distance is utilized; the details of this distance metric are explained in Section III.

The representation of the energy function in terms of the previous energy function is defined separately for $U$ and $V$ terms. For the unary terms ($U$), the unary energy of each region in the current frame $t$ can be obtained by the weighted sum of the energy of each region in the previous frame $t-1$. Moreover, these weights can be selected inversely proportional to these distances between regions. In other words, for any region in time $t$, unary term is written in terms of the weighted sum of the unary terms in time $t-1$ as;
\begin{equation}
\label{eq1}
U^t(\alpha ^t_i, z^t_i) = \frac{1}{\gamma ^t_i} \sum_{v^{t-1}_j \in \mathcal{V}^{t-1}} U^{t-1}(\alpha_j^{t-1}, z^{t-1}_j) e^{-dis(z^t_i,z^{t-1}_j)}
\end{equation}
where superscripts represent the time instants and $dis(z^t_i,z^{t-1}_j)$ represents the aforementioned distance metric between region $i$ in time $t$ and region $j$ in time $t-1$. $\gamma ^t_i$ is used for normalization and it can be computed as $\gamma ^t_i = \sum_{v^{t-1}_j \in \mathcal{V}^{t-1}} e^{-dis(z^t_i,z^{t-1}_j)}$.

The main rationale behind the relation in (\ref{eq1}) can be explained by considering each region in the previous frame as a model, and assuming the current setup as a mixture of these already calculated models. As in the case of mixture models, the resultant unary energy is a weighted sum of the each model in the mixture. Another rationale can be put forward by considering each region in the previous frame as a moving object. Therefore, every region in the current frame will either corresponds to an object in the previous frame or alpha matting of different overlapping objects. When the small segments generated by an over-segmentation algorithm or even pixels are considered as regions, every region in the current frame should have many matches in the previous frame; therefore, instead of exact matches, linear combination of all matches can be used. The selection of these matches and computation of their corresponding weights are implicitly performed by the help of the proposed distance metric.
 
For the binary terms ($V$), the conventional approach is using the frequently utilized Potts Model; {$ V(z_i,z_j) = exp^{-\frac{|z_i-z_j|}{\beta}}$} ($\beta$ is used for normalization). This penalty is generally considered as inversely proportional to the color differences in order to generate coherent segmentation results. However, if we use this conventional approach, it is impossible to use the proposed dynamic optimization method explained in Section IV because both unary and binary energy terms of current frame are assumed to be linearly dependant on unary and binary energy terms of previous frame. Therefore, we slightly change the coherence penalty; we consider the graph of edges (i.e. dual graph of nodes) and apply the same propagation rule to this graph. Therefore, the relation between the current and previous binary energies are represented as follows:
\begin{equation}
\begin{split}
V^t(z^t_i, z^t_j) = \frac{1}{\gamma ^t_{ij}} &\sum_{v_k \in \mathcal{V}^{t-1}} \sum_{v_l\in N(v_k)} e^{-dis(z^t_i,z^{t-1}_k)} . \\&
e^{-dis(z^t_j,z^{t-1}_l)} V^{t-1}(z^{t-1}_k,z^{t-1}_l)
\end{split}
\end{equation}
This definition of the binary energies can be related to smoothing the conventional binary terms. For the case of linear and piecewise linear penalty functions, one can show that the definition of the proposed binary penalty is equivalent to applying spatio-temporal edge-aware smoothing to video frames; then, computing conventional binary terms. In other words, defined binary energy is equivalent to the conventional binary energy used in \cite{boykovMethod} with an extra prior smoothing operation. Hence, defined binary energy can be considered as color smoothness penalty with an additional temporal smoothness.

Main advantages of the proposed estimation method are generality in terms of distance function selection and extended support region. Although the proposed system is based on geodesic distances, the proposed energy propagation can also be accomplished by a variety of distance functions accompanying color, motion and shape. Moreover, in this formulation, each region in the previous frame actually supports every region in the current frame. Therefore, any region is supported by all the regions of the previous frame. Such an approach relaxes limited motion assumption significantly, and eliminates the necessity to use any global model.

\begin{figure}[h!]
\vspace{-3mm}
        \centering

        \subfloat[ $U^{1}(\alpha_i^{1},z_i^{1})$]{\includegraphics[width=0.23\textwidth]{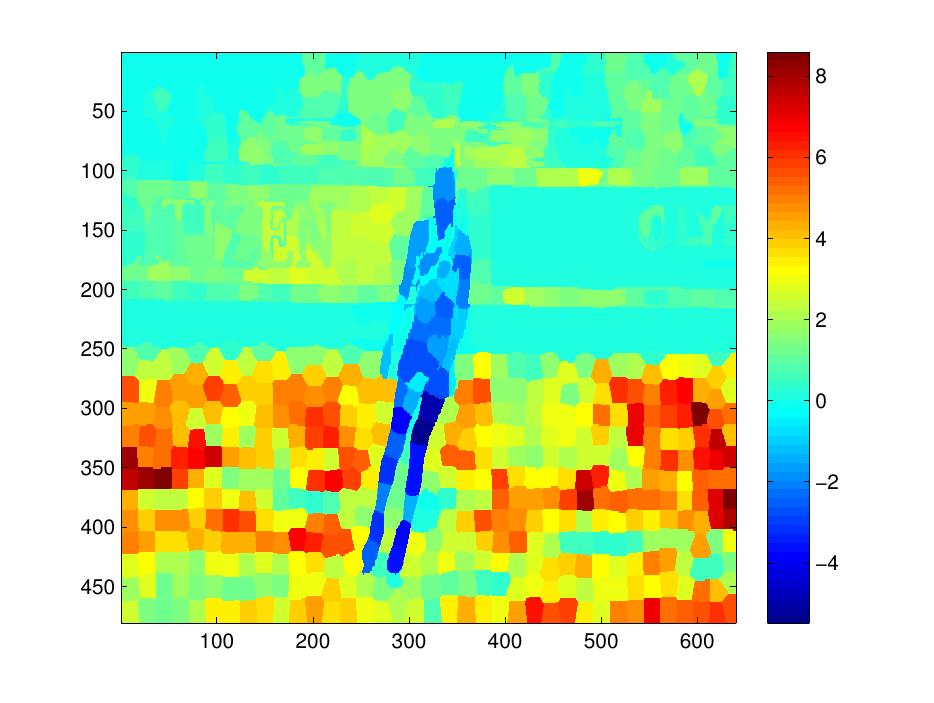}}
		\subfloat[$V^{1}(z_i^{1},z_j^{1})$]{\includegraphics[width=0.23\textwidth]{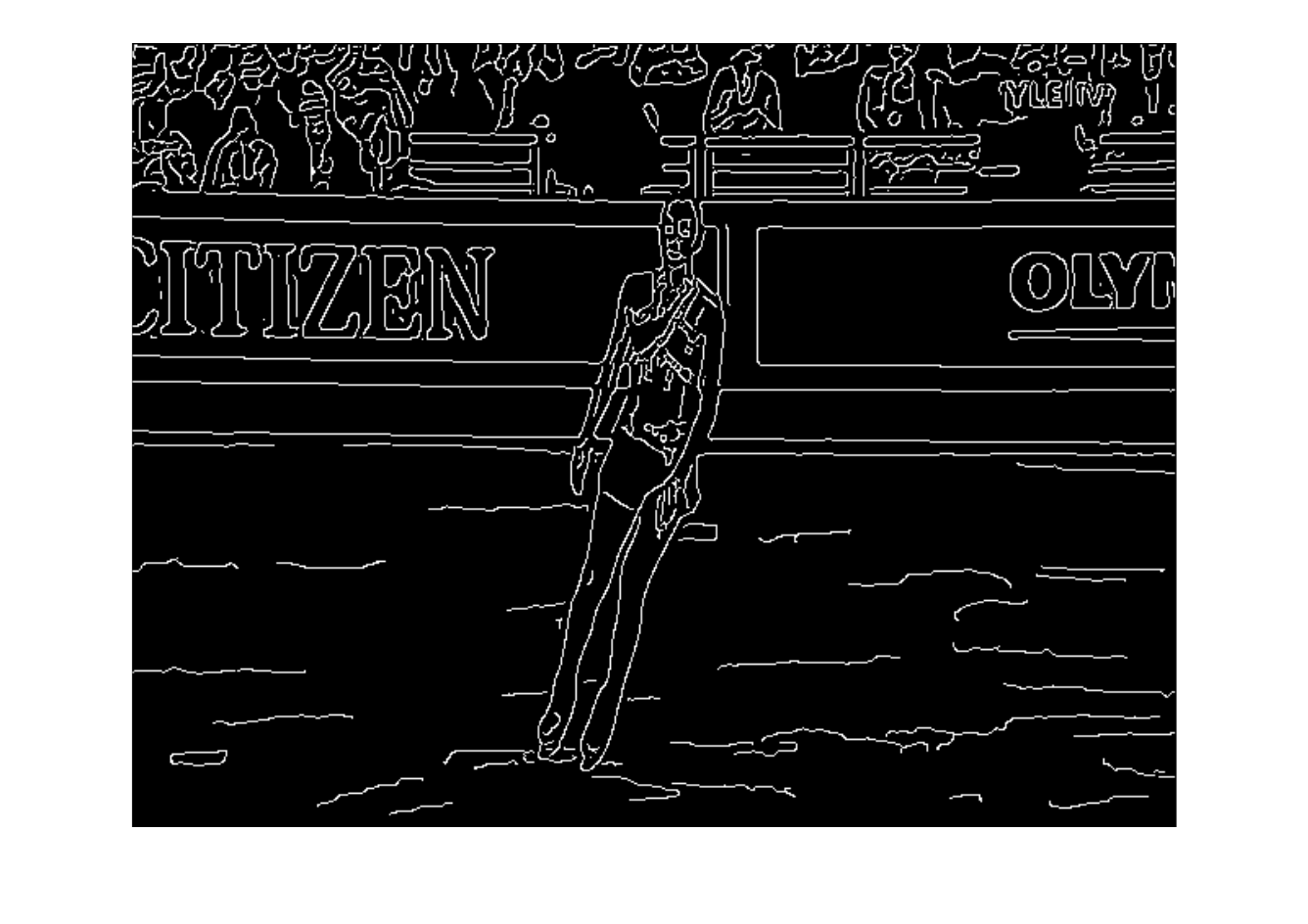}}\\
		 \vspace{-4mm}
		\subfloat[ $\hat{U}^{5}(\alpha_i^{5},z_i^{5})$]{\includegraphics[width=0.23\textwidth]{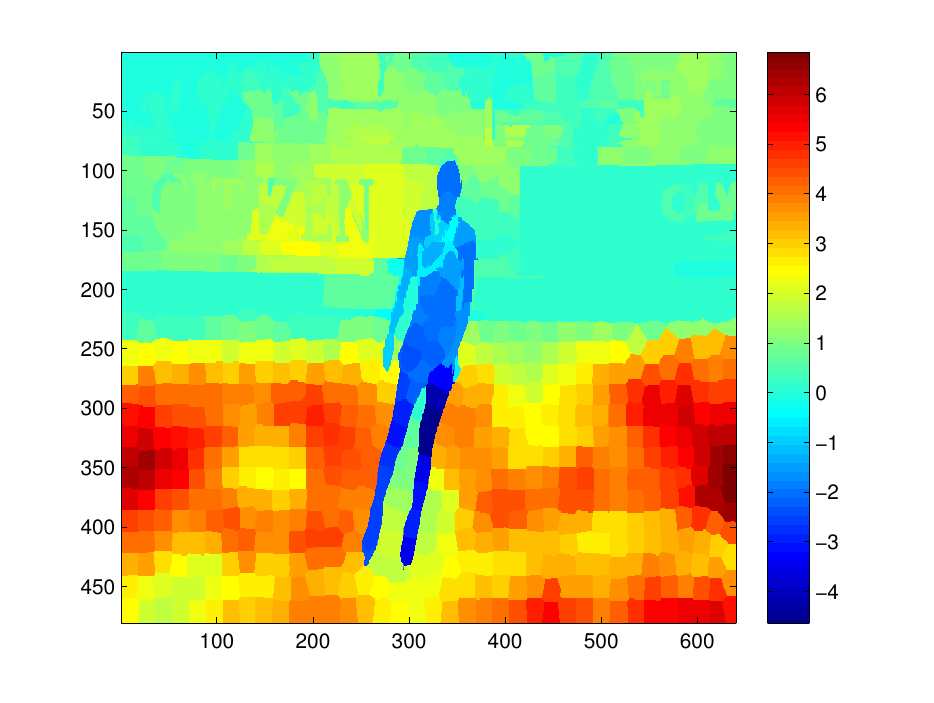}}
		\subfloat[$\hat{V}^{5}(z_i^{5},z_j^{5})$]{\includegraphics[width=0.23\textwidth]{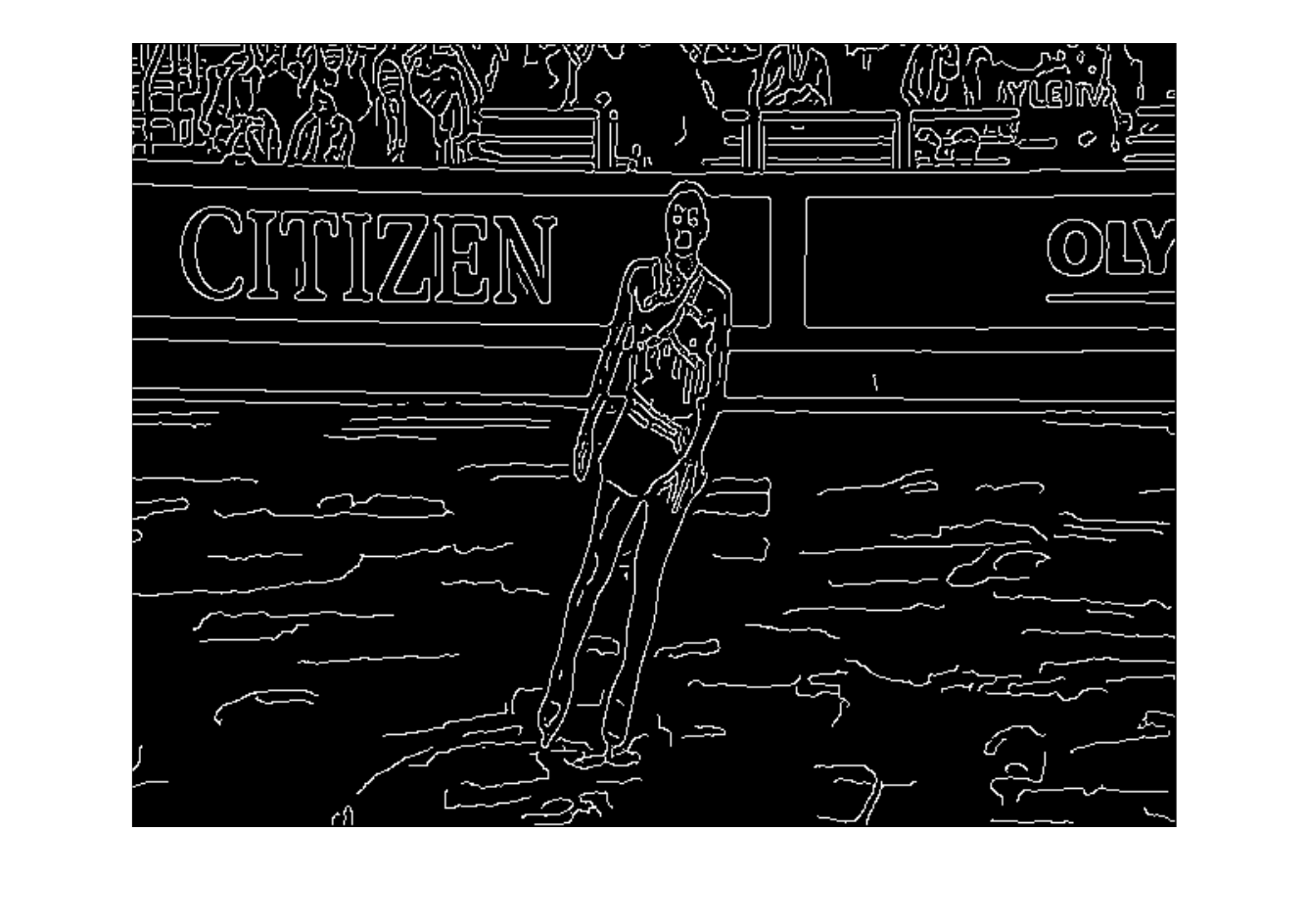}}
		
        \caption{Unary $U^t(\alpha ^t_i, z^t_i)$ (a) and Binary energy $V^t(z^t_i, z^t_j)$ (b) terms of frame t=1 is computed via user interaction, whereas energy terms of frame t=5 is estimated via proposed energy propagation method. Estimated unary energy (c) is consistent with the actual foreground/background probabilities of the regions. Furthermore, estimated unary energy (c) is much smoother than the unary energy found via interaction (a) due to the wide support region.}\label{fig:prop}
        \vspace{-1mm}
\end{figure}
In order to visualize energy propagation, we apply the proposed method to a typical example. The energy of the first frame is computed via the interactive image segmentation method \cite{mmpd}, and the energy of the $5^{th}$ frame is estimated via the proposed method. As in the case of \cite{mmpd}, over-segments are obtained by using SLIC algorithm \cite{slic}. The resultant propagation is illustrated in Figure \ref{fig:prop}; where it suggests that the estimated unary terms are much smoother than the original terms due to the wide support region. It can also be observed that utilization of (2) for binary weights is feasible and the result is consistent with the edge map. 

\section{Efficient Computation of Bilateral Filters - IP/BE Filter}
Energy propagation method explained in Section II, estimates the MRF energy of each frame. Hence, segmentation results can be obtained via minimization of these energies. However, the proposed energy propagation method is not applicable to interactive and mobile multimedia applications due to high computational complexity of bilateral filters and computation of geodesics. It should be noted that the relations in (1) and (2) correspond to a cross filtering by a bilateral filter \cite{cevo}. Naive implementation of a bilateral filter requires $O(n^2)$ operation; indeed, each operation requires a computation of a geodesic distance which has a linear time complexity via fast marching  algorithm \cite{march}. Therefore, the overall complexity of the estimation is $O(n^3)$ which is far from being acceptable. In order to solve this efficiency problem, we utilize the recently proposed approximation of geodesic bilateral filter \cite{cevo, bxp}.

Bilateral filters are widely used for edge-preserving smoothing in computer vision \cite{bl1}. They can be considered as smoothing filters, whose weights are obtained from a Gaussian kernel. Later, bilateral filters are also attributed to anisotropic-diffusion filters \cite{ad}. However, their high computation cost make them unfavourable until some recent improvements over the computation time. In \cite{hdr}, bilateral filter is accelerated by using quantization and piecewise linear approximations. Alternatively, in \cite{linconv}, bilateral filtering is performed by using linear convolutions in higher dimensions. Recently, a constant time bilateral filter is proposed independently by Cigla et. al. \cite{cevo} and Thevenaz et. al. \cite{bxp} under different names as \emph{Information Permeability (IP)} and \emph{Bi-Exponential (BE) filter}. For the sake of fairness, we refer to this filter as IP/BE.

In IP/BE filter, bilateral filtering is performed by using a dual one-tap recursive filter (IIR) for each dimension in the 2D spatial domain (vertical and horizontal). By the help of the recursive filter, one can achieve full image support for any pixel by a Gaussian kernel. 1D one tap recursion filter is applied in two directions (both horizontal and vertical) separately. Consider the sequence $x[n]$ of length $N$, the recursion is applied as follows;

For positive direction (left to right, \cite{cevo, bxp}):
\[
\hat{x}_1[k] = x_1[k] + \hat{x}_1[k-1] r(x[k],x[k-1]) 
\]
whereas, for negative direction (right to left):
\[
\hat{x}_2[k] = x_2[k] + \hat{x}_2[k+1] r(x[k],x[k+1]) 
\]
In this representation, $r(x[k],x[k-1])$ is the filter weight and can be computed as $r=exp(-\frac{|x[k]-x[k-1]|}{\sigma^2})$. Moreover, this recursion should be initialized as $x_1[n]=x_2[n]=x[n]$. The values after recursion needs an extra normalization operation. During this normalization, the constants are computed via the same recursion. Consider a sequence of 1's with length $N$, as $1_1[n]$ and $1_2[n]$, then the normalization constants are, for positive direction:
\[
\hat{1}_1[k] = 1_1[k] + \hat{1}_1[k-1] r(x[k],x[k-1]) 
\]
and, for negative direction:
\[
\hat{1}_2[k] = 1_2[k] + \hat{1}_2[k+1] r(x[k],x[k+1]) 
\]
The final smoothed values are computed as \cite{cevo, bxp}:
\[
y[k]=\frac{\hat{x}_1[k]+\hat{x}_2[k]}{\hat{1}_1[k]+\hat{1}_2[k]}
\]
These 1D recursion filters can be applied to horizontal, spatial and temporal directions separately. Moreover, these results can be combined as explained in \cite{cevo,bxp}. It should be noted that, IP/BE filter is explained for smoothing scenario. In the proposed method, IP/BE filter is used as a cross filter. $x[k]$ values corresponds to unary and binary energy terms at $t-1$ and, $\hat{x}[k]$ values correspond to unary and binary energy terms at $t$. Moreover, the weight values are obtained via color values of the image instead of the energy terms. In other words, $r(x[k],x[k-1])$ is replaced with $r(z_k,z_{k-1})$. It should also be noted that these 1D filters are applied on horizontal, vertical and temporal dimensions separately. Hence, their ordering changes the result. We apply the filters in both orders in the spatial domain, then in the temporal domain (horizontal, vertical, temporal and vertical, horizontal, temporal), and use the average of the results as the final propagated energy.

In \cite{cevo}, it is argued that IP/BE filter approximates a geodesic bilateral filter. Although the weights do not correspond to the geodesic distances, they correspond to the cost of the path having a single horizontal and a single vertical piece. Moreover, the quality of this approximation is demonstrated in an edge preserving smoothing problem \cite{cevo,bxp}. Although this approximation could have some failure cases, it can still be used within (1) and (2) in most of the practical cases.

\begin{figure}[h!]
\vspace{-2mm}
        \centering
		\includegraphics[width=0.4\textwidth]{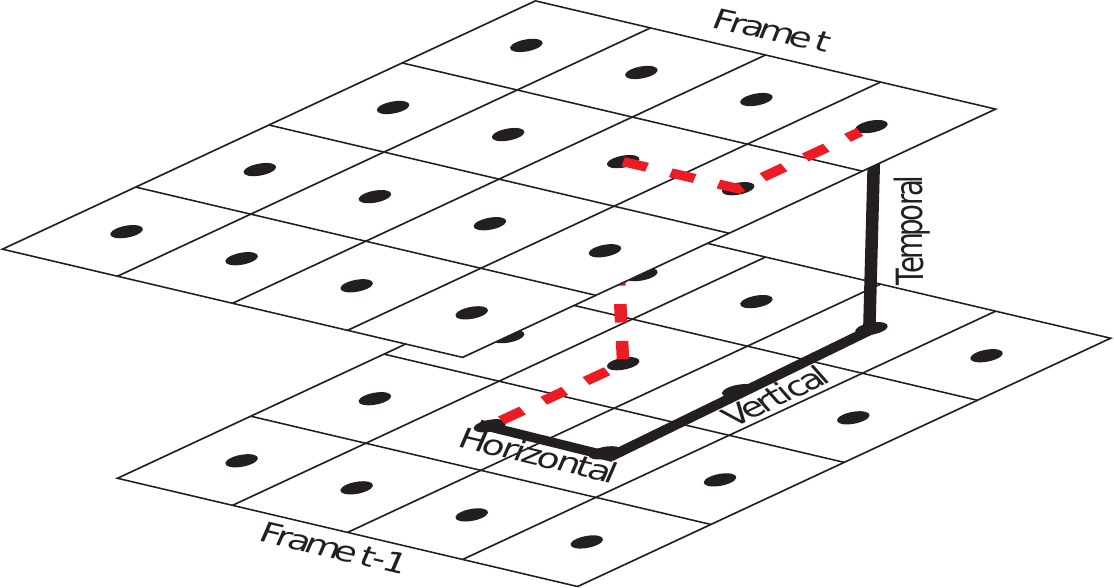}
		\caption{Illustration of step-like paths used in IP/BE filter. Instead of a geodesic path shown as red dashed line (arbitrarily selected), a path having single horizontal, vertical and temporal components is used.}
		\vspace{-2mm}
		\label{bx}
\end{figure}

When such filtering is used, the resulting MRF energy for the current frame is equivalent to the energy defined in Section 2 by $dis(z_i^t,z_j^{t-1})$ as the cost of the step like path shown in Figure 2. In other words, instead of taking the distance as the minimum of the sum of color differences among the paths connecting point $z_i^t$ and $z_j^{t-1}$, this distance is taken as the sum of color differences along the path having three components on each dimension ($x,y$ and $t$). For the videos not having serious level of noise or high amount of motion, the proposed cost approximates the geodesic distance \cite{cevo}. 

When the proposed method is analysed, the number of operations is linear with the number of regions in the frames, if the number of overlapping regions is bounded by a constant. Indeed, most of the over-segmentation algorithms has a parameter which control the number of superpixels or the maximum size of a superpixel. Therefore, maximum number of overlapping regions can be adjusted. Hence, by proper selection of over-segmentation parameters, the proposed method runs in linear time. In summary; it is possible to approximate MRF energy propagation explained in (1) and (2) via algorithm having linear time complexity by using the IP/BE filter.

\section{Bilateral Dynamic Graph-Cut}
We propose a linear-time method to estimate MRF energy of a frame by using the previous frame. Moreover, for each frame, this energy function can be minimized via min-cut/max-flow method optimally \cite{boykovMethod}. However, due to our challenging efficiency requirements, we need to further improve min-cut/max-flow approach to increase the overall efficiency. Hence, we propose a method to recycle residual flows in a bilateral filtering scenario for non-lattice graphs.

In general, MRF energy minimization problem is converted to min-cut problem on two terminal graph as an initial step. In the constructed graph, there are two terminals, namely source and sink (i.e. foreground and background) and every node is connected to these terminals by the terminal edge weights equal to unary energies. Moreover, non-terminal edge weights are selected as the binary energy terms. One can show that finding minimum cut on this graph is equivalent to minimizing the MRF represented energy. It is also shown that \cite{network} finding minimum cuts is equivalent to finding maximum flows which can be pushed from source to sink. 

Typically, the solution to the max-flow problem is obtained via augment paths algorithm \cite{boykovMethod}. This algorithm can be explained by using the residual capacities and augmenting paths. Residual capacity $r_{ij}$ of the edge $(i,j) \in E$ is the maximum additional flow that can be sent from $i$ to $j$ through edge $(i, j)$. The augmenting path is the path from source to sink through unsaturated residual edges (edges with positive residual weights). Augmenting paths algorithm uses the fact that pushing any flow through an augmenting path does not change the solution \cite{boykovMethod}. In other words, the solution to the original graph $G$, and the graph $G'$ which results from pushing a flow through an augmenting path is equivalent. The algorithm iterates until there exist no augmenting path in the graph; then, min-cut is the cut through edges having 0 residual weight. Since the cost of this cut is equal to zero, by sub-modularity, there can not be better solution.

On the other hand, solving each frame separately is redundant. Augmenting flows in the previous frame can actually be "recycled" for the next frame. There exist a dynamic method \cite{dynGCut} to solve graph-cut for each frame dynamically. In this dynamic algorithm \cite{dynGCut}, the structure of the graph assumed to be not changing throughout the video, and the edge weights change slightly throughout iterations. Therefore, if the residual graph of the previous iteration can be used, the computational burden might significantly decrease. In \cite{dynGCut}, a method for this edge update is developed. If a weight $w^{t-1}_{ij}$ of edge $(i,j)$ is changed to $w^{t}_{ij}$ , than the solution to this new graph can be determined by solving the updated residual graph by the update $r^{t}_{ij} = r^{t-1}_{ij} + w^{t}_{ij} - w^{t-1}_{ij}$. Indeed, this relation corresponds to pushing same flows through same nodes in the next frame. Only possible problem is negative edge weights and it is already solved in \cite{dynGCut}.

However, this method is not applicable to a general case, since the structure of the graph might change significantly due to superpixel-based representation. Over-segment positions and sizes might change in each iteration of the over-segmentation method. Therefore, we propose another dynamic version of the min-cut/max-flow algorithm for such a varying graph structure case in linear filtering scenario.

In order to apply the conventional dynamic method to varying graph structure case, a computationally expensive graph matching problem need to be solved between graph of the current frame and that of the previous frame. However, in the proposed method, there exist a linear relation between each node in the current frame and the previous frame. Moreover, edge weights of the graph of the current frame is defined in terms of the edge weights of the graph of the previous frame, as in (1) and (2). Therefore, this relation needs to be exploited. 

\begin{figure}[ht]
	\vspace{-3mm}
        \centering
		\includegraphics[width=0.5\textwidth]{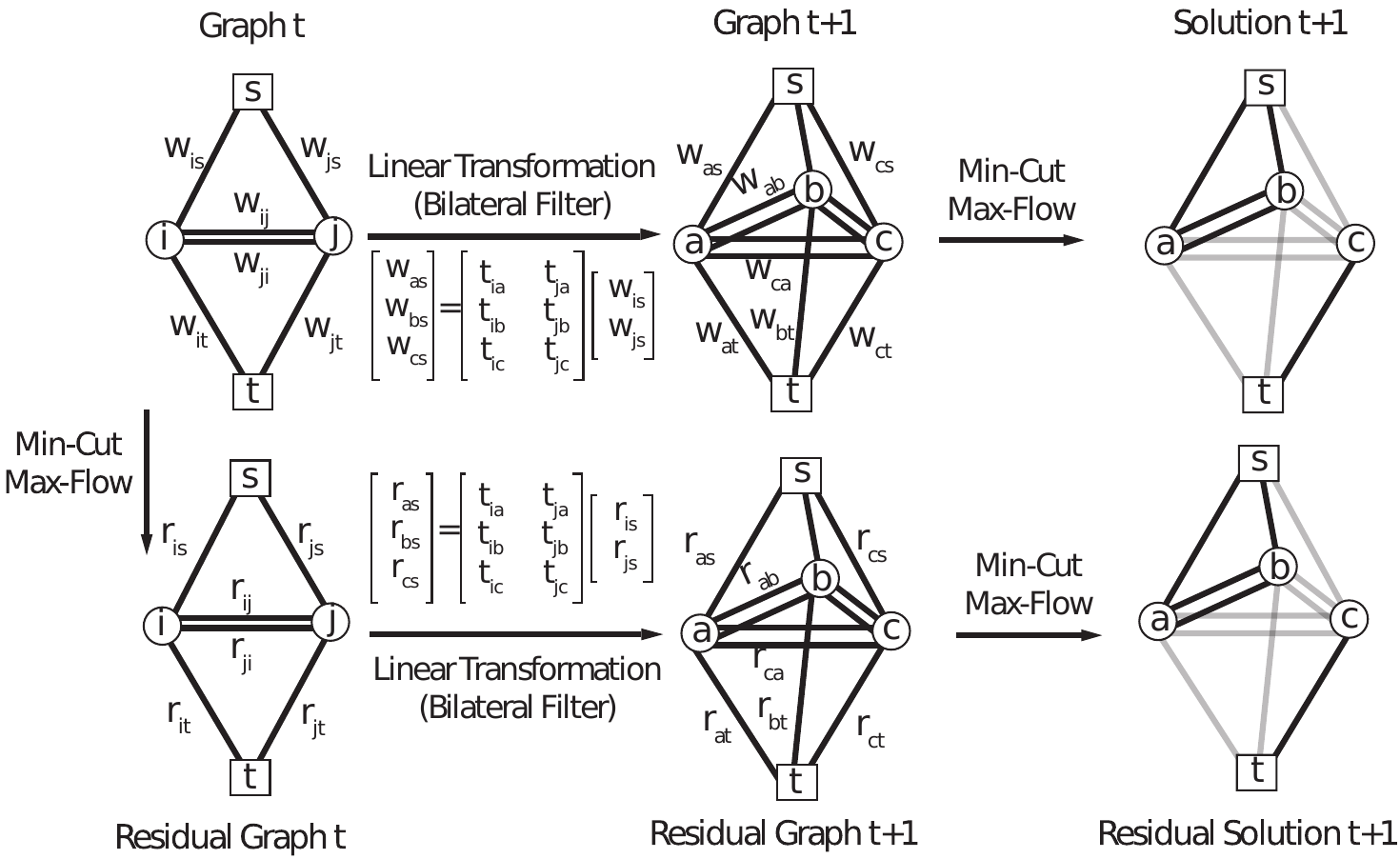}
		\caption{{\bf Example for Proposition 1.} The solution at $t+1$ is obtained by applying linear transformation to the graph at $t$ and solving graph at $t+1$. On the other hand, residual graph at $t+1$ is obtained by applying same linear transformation to the residual graph at $t$ which is already computed while solving graph at $t$. Moreover, solution for $t+1$ is obtained by solving residual graph at $t+1$. Proposition states that the solution at $t+1$ and residual solution at $t+1$ are equal. (In order to make the illustration less dense $w_{ba}, w_{bs}, w_{ac}, w_{bc} $, $w_{cb}, r_{ba}, r_{bs}, r_{ac}, r_{bc} $ and $r_{cb}$ are not shown in the figure).}
		\vspace{-4mm}
		\label{proof}
\end{figure}

We propose to propagate the flows in the previous frame to the current frame via bilateral filter computed during the estimation step of MRF energies. More interestingly, we show that if the flows in the previous frame are propagated and pushed to the current frame, the resultant residual graph will be equivalent to applying the same bilateral filter to the residual graph of the previous frame. In other words, in order to find the updated residual graph in time $t$, we simply apply the proposed bilateral filter to the residual graph in time $t-1$. Moreover, we claim that minimum cut result will be same. This claim is also explained in Figure \ref{proof} by an example. In order to keep the results as general as possible, we prove this claim for the general bilateral filtering scenario.

\begin{proposition}
\label{prop}
Binary labels obtained by minimizing the MRF energy, resulted after applying bilateral filter on the energy function which is defined via residual graph, is equivalent to minimizing the MRF energy obtained via applying bilateral filter on the original energy function.
\end{proposition}

The proof of this proposition is deferred to Appendix. By using Proposition 1, instead of propagating MRF energy completely, we only propagate residual graphs throughout the frames. This dynamic propagation results in significant computation time decrease as explained in Section 5.2. Efficiency obtained via dynamic graph-cut can be crucial factor to increase the interaction quality for interactive multimedia applications. On the other hand, the proposed technique is the only dynamic min-cut/max-flow method which is applicable to the graphs whose structures are varying.  

In summary, the proposed video segmentation method starts with an already known MRF energy for the first frame, solves the min-cut/max-flow problem; then, propagates the residual energy to the next frame via the proposed linear time algorithm and continues to iterate through frames.
 
\section{Experimental Results} 
As explained in Section I, the proposed MRF energy propagation can either be used as an interactive video segmentation algorithm with an existing interactive image segmentation method or as a speed-up method for an automatic video segmentation algorithm. Hence, we have experimented the proposed method for both scenarios. In Section V.A, we utilize recently proposed mobile and dynamic image segmentation tool \cite{mmpd} in order to experiment the interactive video segmentation scenario. Moreover, segmentation quality and computational efficiency results are compared against existing interactive video segmentation algorithms. On the other hand, in Section V.B, we use the proposed method to speed-up a recently proposed automatic video segmentation tool \cite{keysegments}. We compare segmentation quality and computational efficiency of the original method and proposed speed-up extension.  

Within the experimental procedures, we tried to keep effect of parameters as small as possible. Indeed, only set of parameters which can be tuned are related to either SLIC over-segmentation \cite{slic} or IP/BE Filter. We kept the default parameters of SLIC over-segmentation, and chose 1000 as the number of over-segments. Kernel size of the IP/BE filter is yet another parameter to be chosen. We chose 30 as the kernel size for $R$, $G$ and $B$ kernels. Discussion related to effect of kernel size can be found in \cite{cevo}.
\subsection{Interactive Video Segmentation}
For the evaluation of the interactive video segmentation scenario, we utilize "coloring-based" interactive image segmentation technique \cite{mmpd} to segment the initial frame. Then, we use the proposed method to propagate the initial segmentation results. Throughout the video, only the initial frame is interacted; no other user interaction is applied for the other frames. We compare the proposed method against two high-performing interactive video segmentation methods from the literature \cite{geo2,roto} and one tracking based method \cite{segtrack}. The first one is a geodesic segmentation method \cite{geo2}. Due to the fact that permeability filter \cite{cevo} actually approximate geodesic distances, our proposed method approximates a bilateral filter using geodesic distances as the filter weights. Therefore, our algorithm can be accepted as an extension of \cite{geo2} with a higher support region, additional coherency term and smoother energy propagation. The other algorithm is a local classifier based segmentation method \cite{roto} which is included in Adobe After Effects CS5 \cite{cs5} as the \emph{roto-brush tool}. Motion coherent tracking is one of the state-of-the-art tracking based segmentation methods. We use exactly same interaction for the segmentation of initial frames for all algorithms. Then, without any further interaction, we use these algorithms for entire video sequence. 
\subsubsection{Quality}
\begin{figure*}[t]
        \centering
        \subfloat{\includegraphics[width=0.15\textwidth]{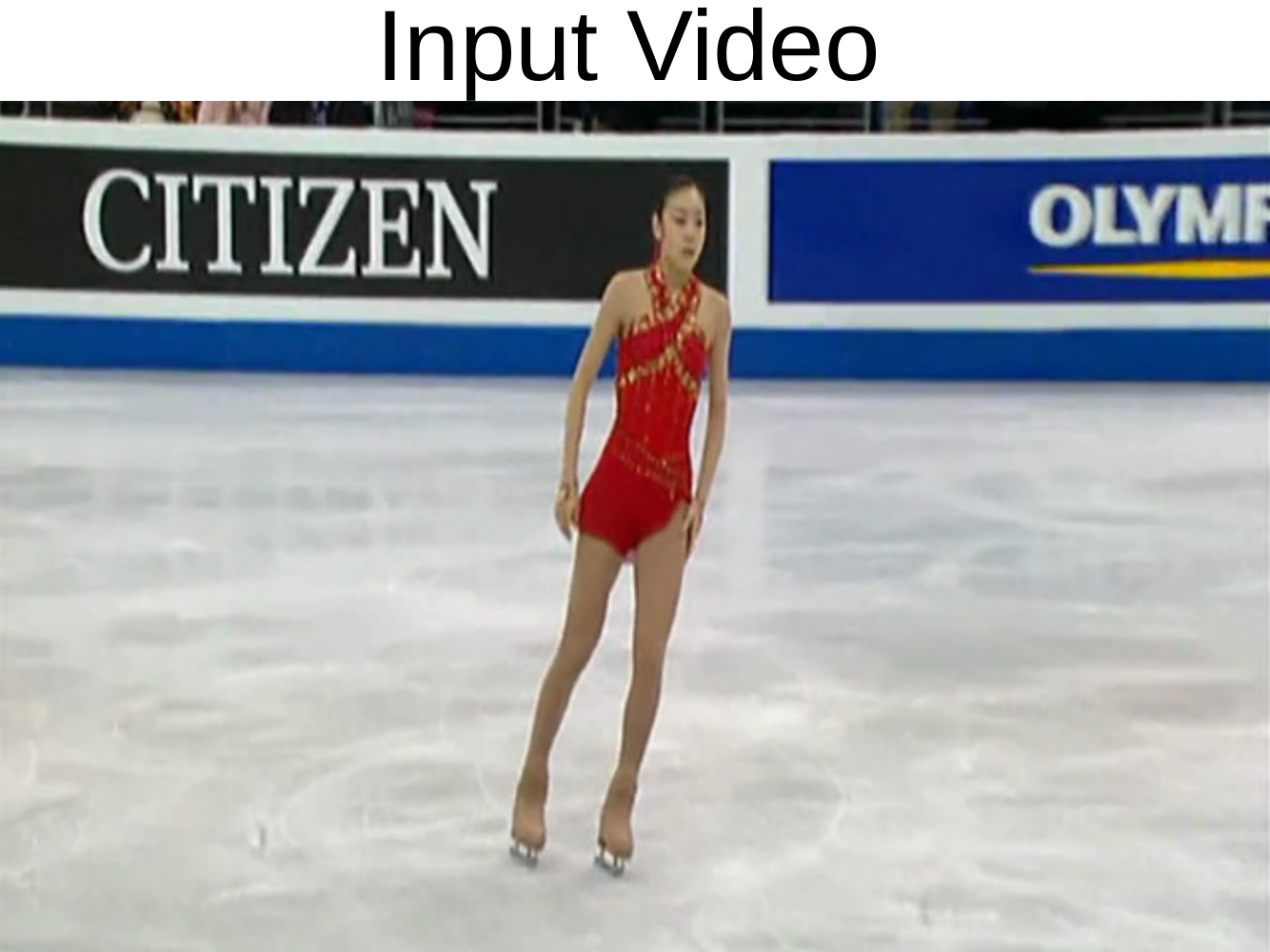}}
        \subfloat{\includegraphics[width=0.15\textwidth]{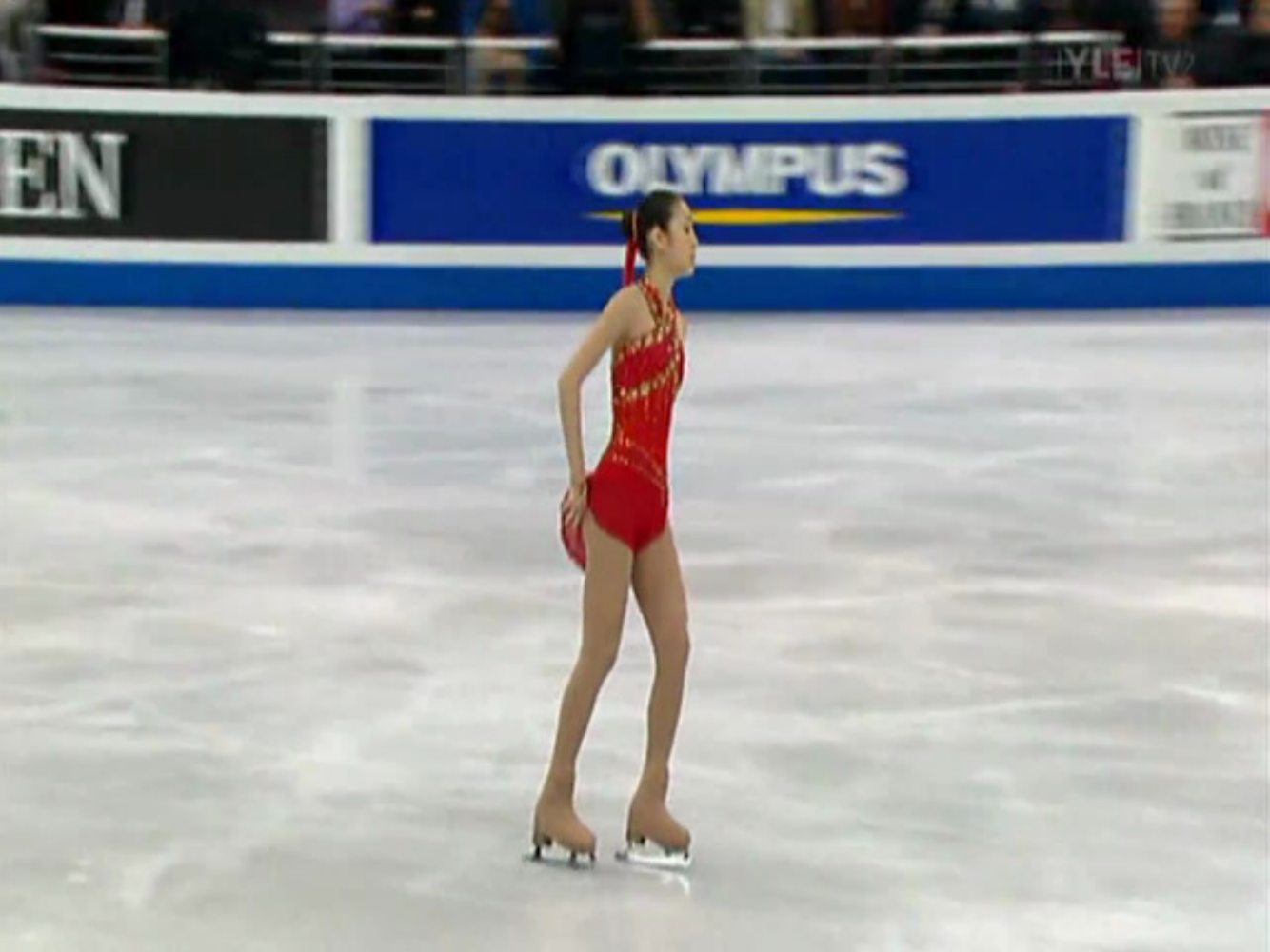}}
        \subfloat{\includegraphics[width=0.15\textwidth]{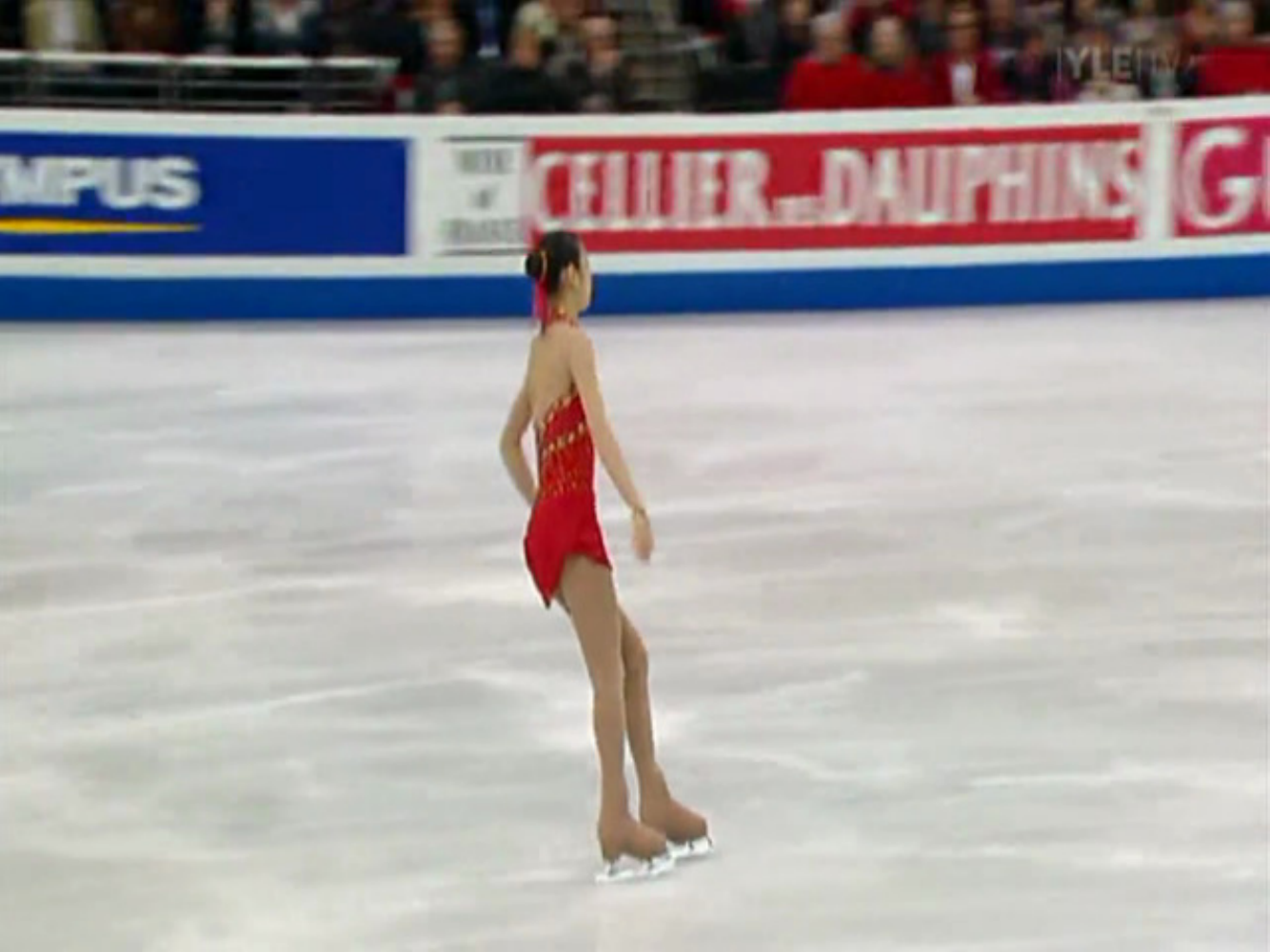}}
        \subfloat{\includegraphics[width=0.15\textwidth]{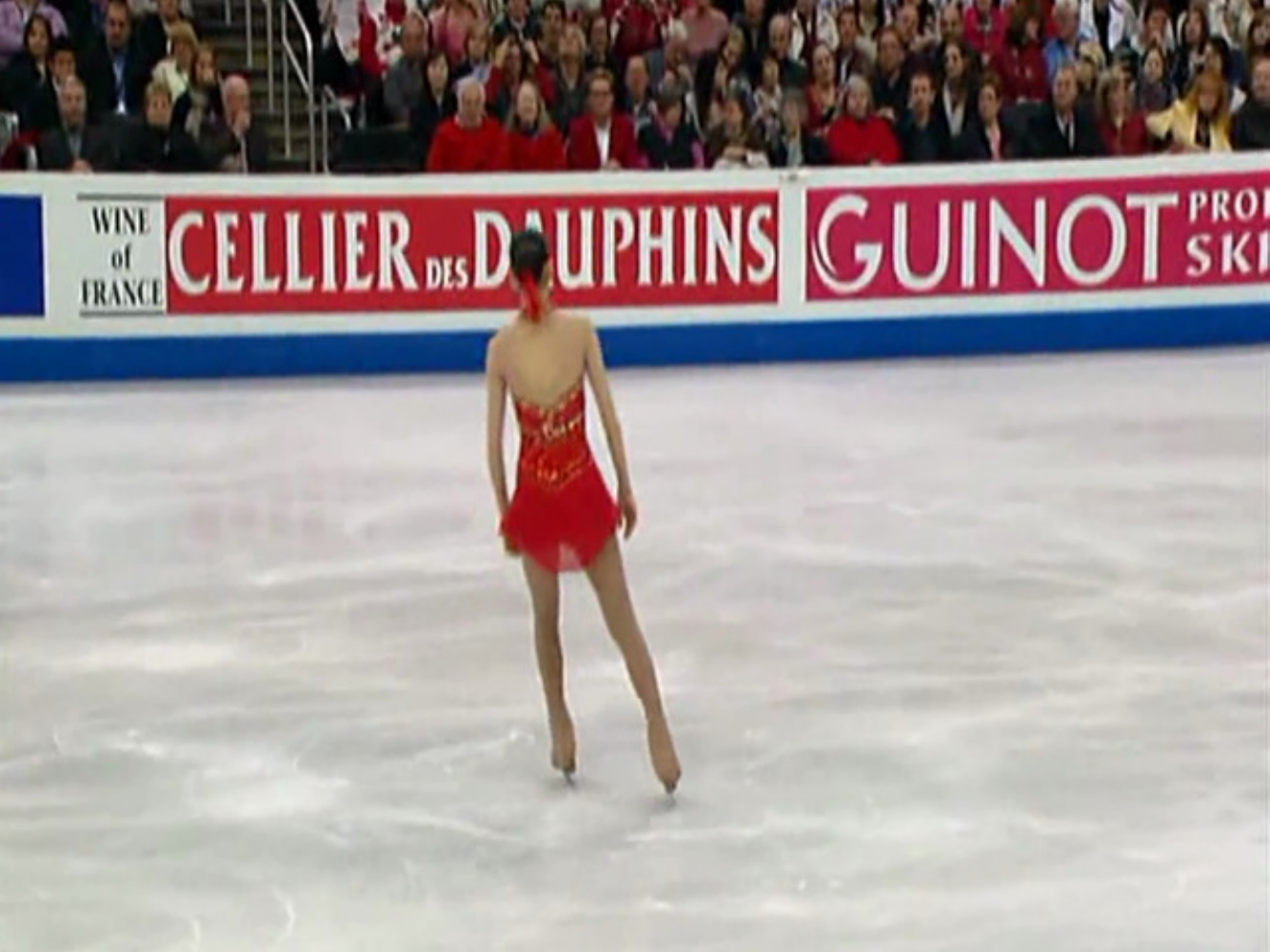}}
        \subfloat{\includegraphics[width=0.15\textwidth]{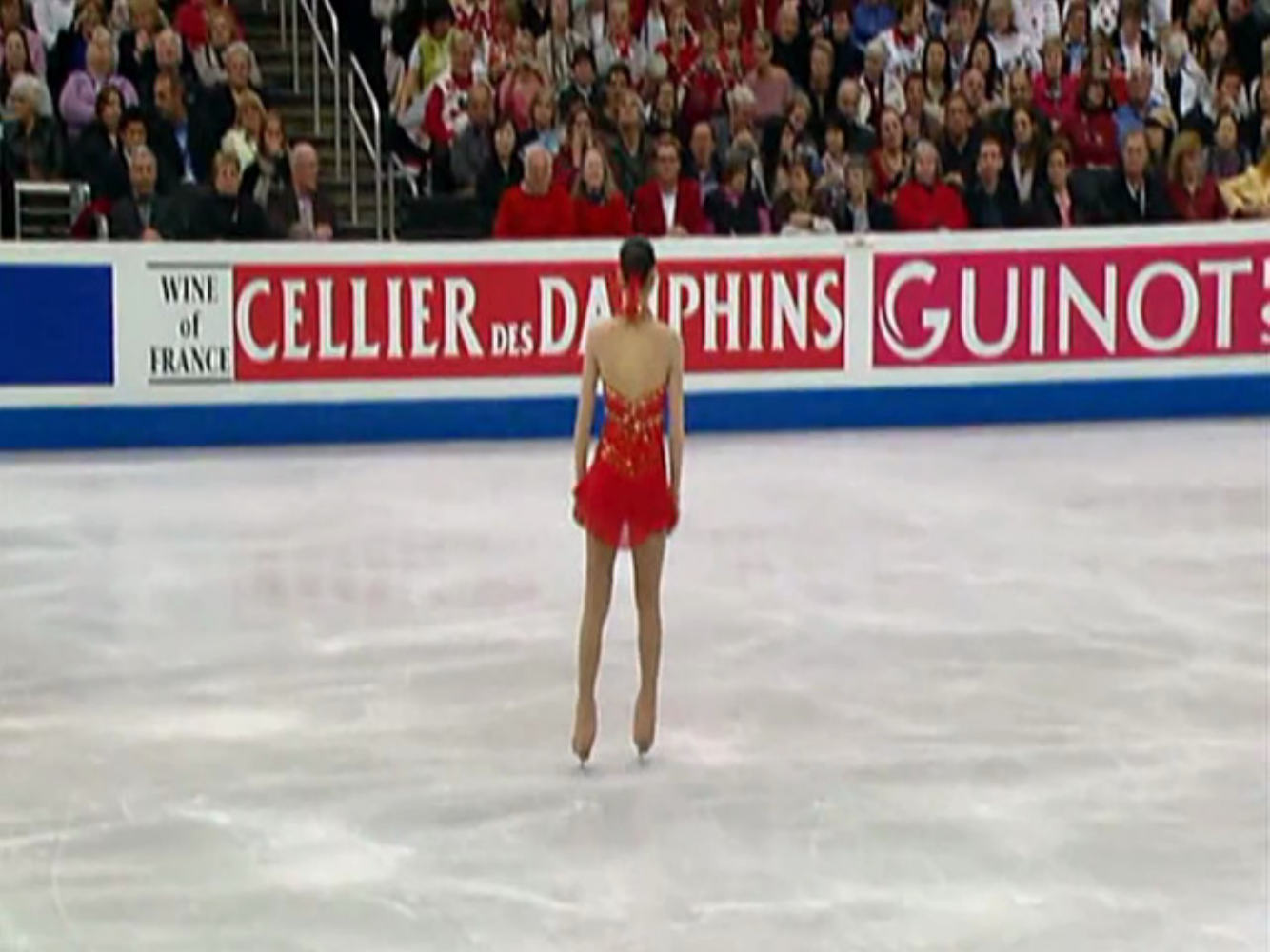}}
        \subfloat{\includegraphics[width=0.15\textwidth]{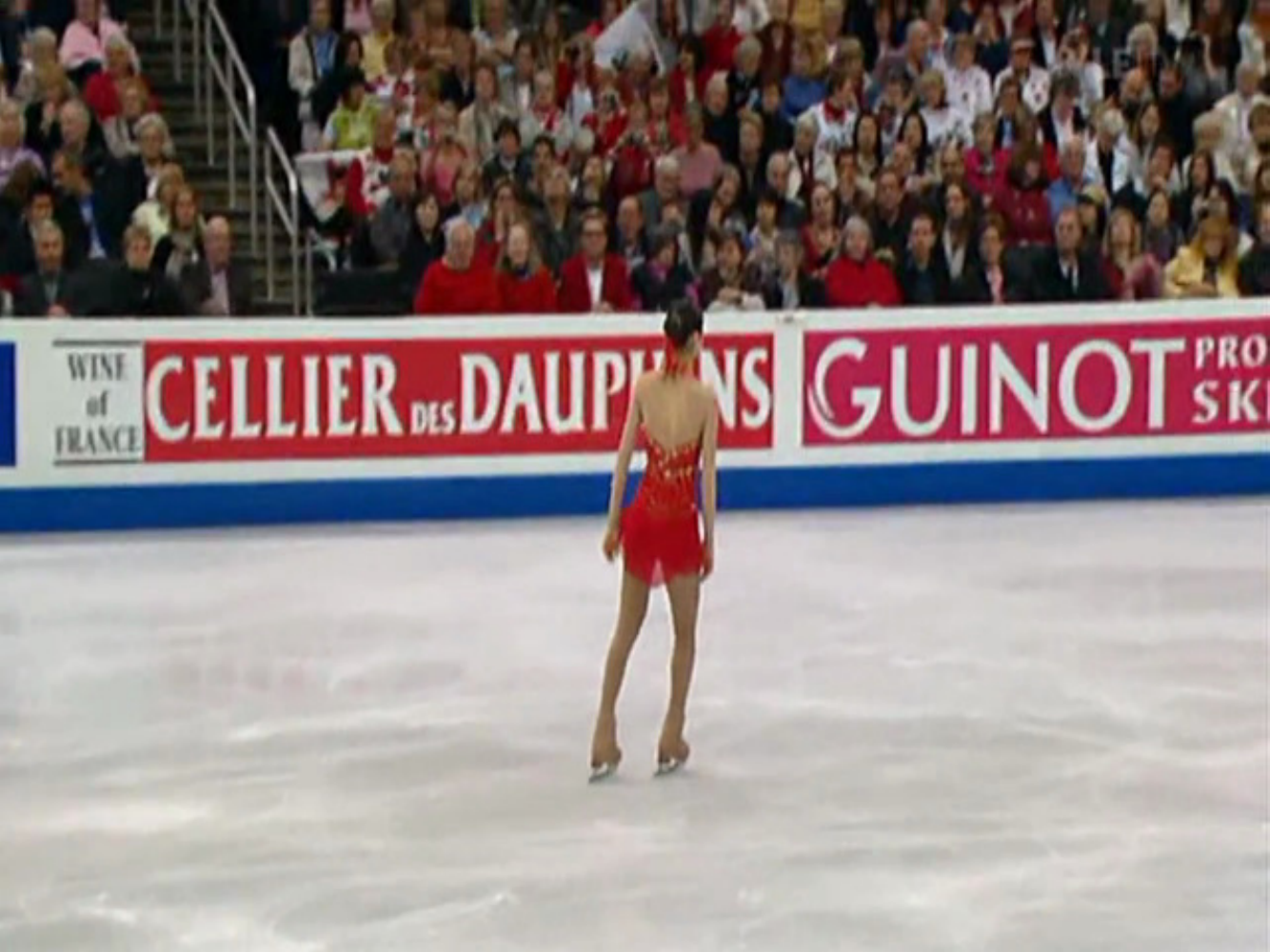}} \\
        \vspace{-3mm}
        \subfloat{\includegraphics[width=0.15\textwidth]{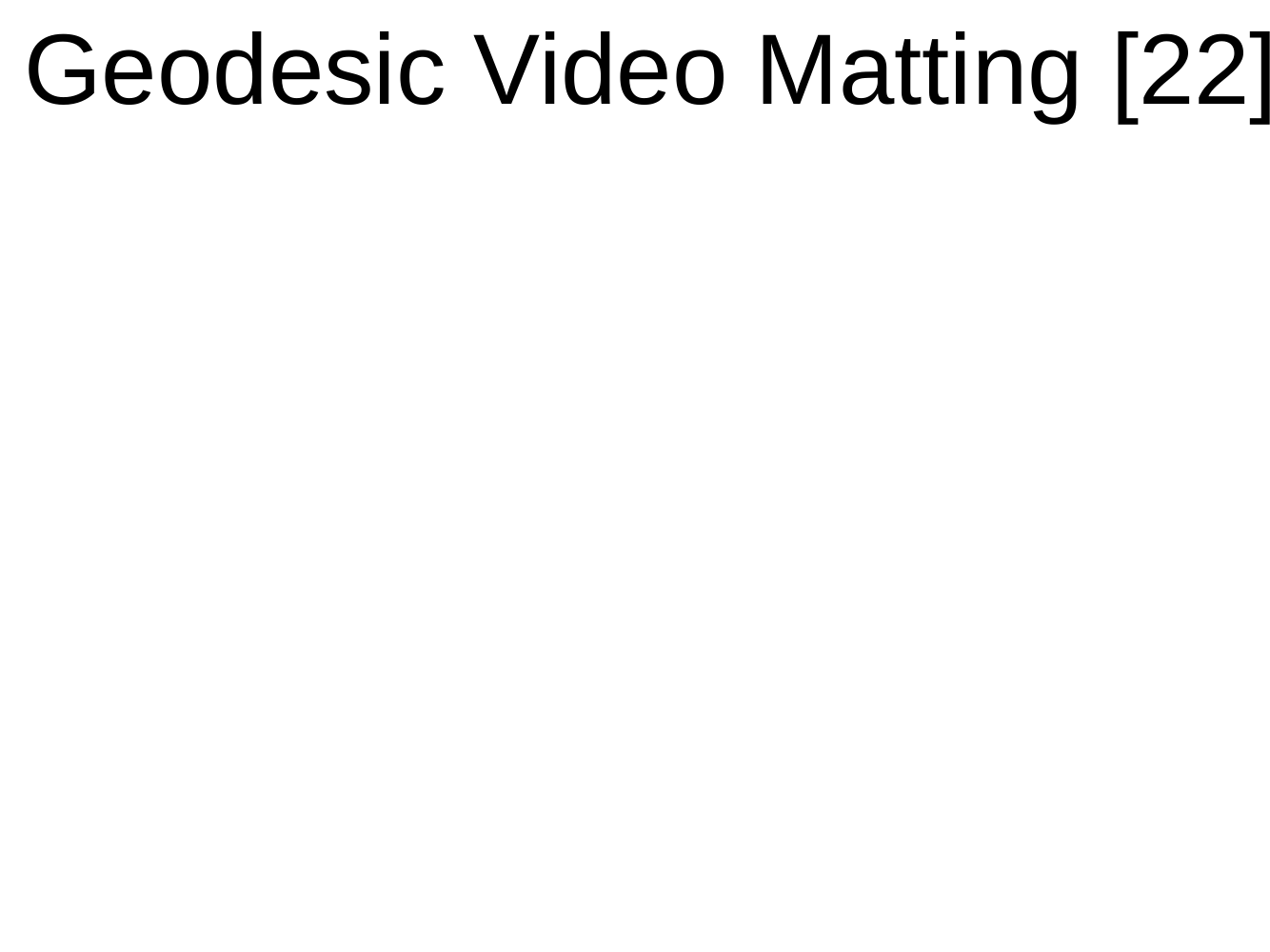}}
        \subfloat{\includegraphics[width=0.15\textwidth]{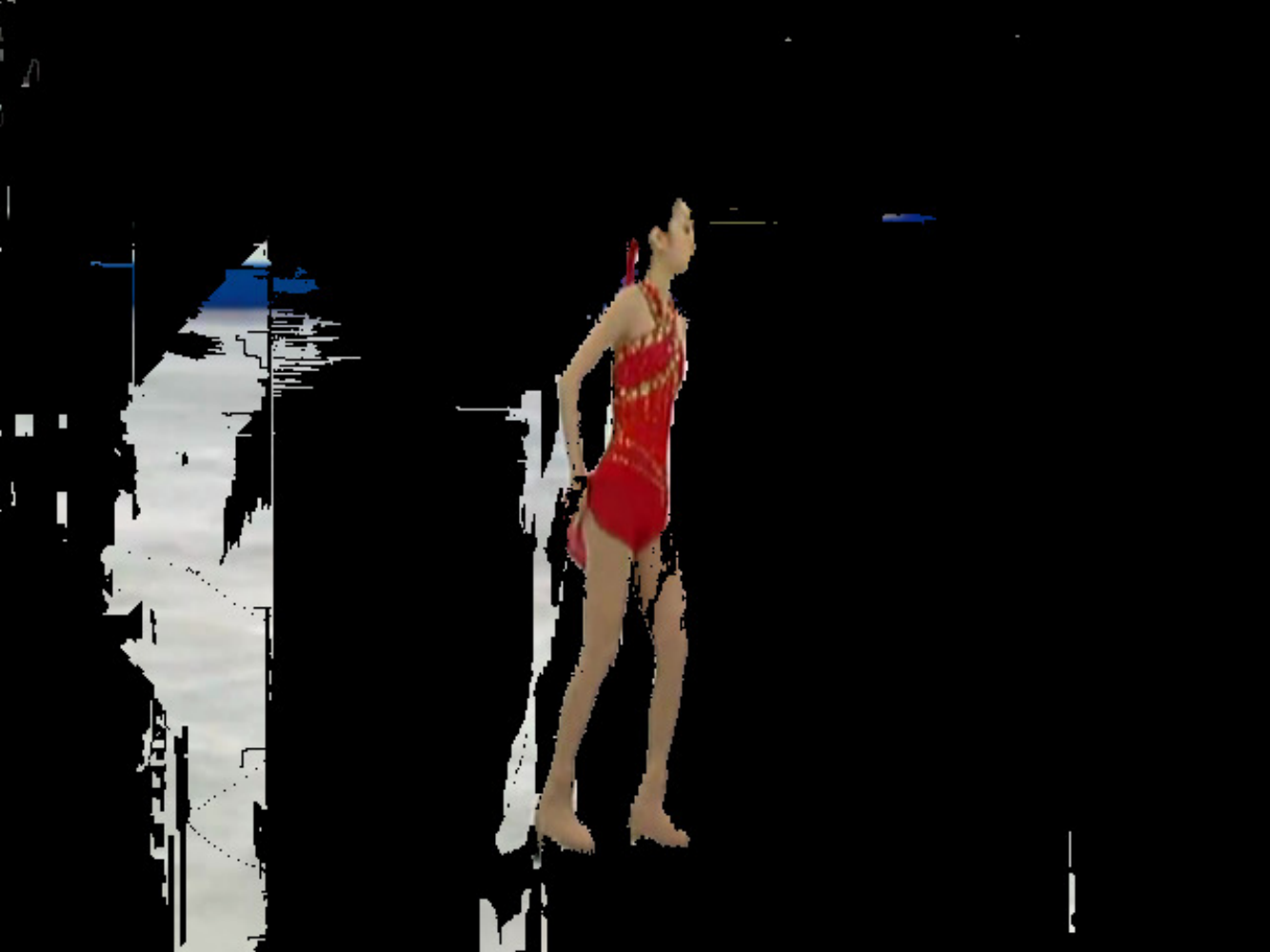}}
        \subfloat{\includegraphics[width=0.15\textwidth]{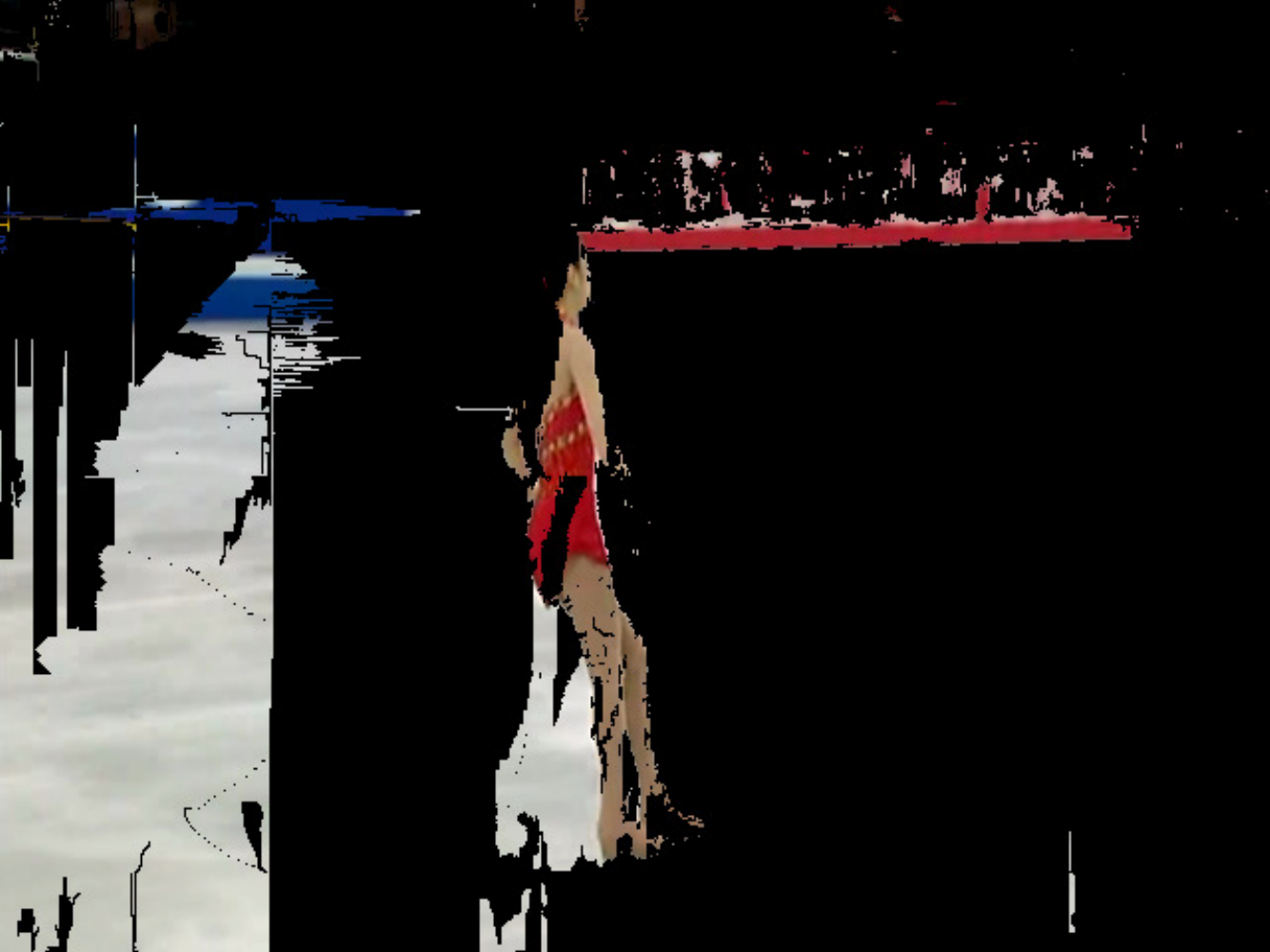}}
        \subfloat{\includegraphics[width=0.15\textwidth]{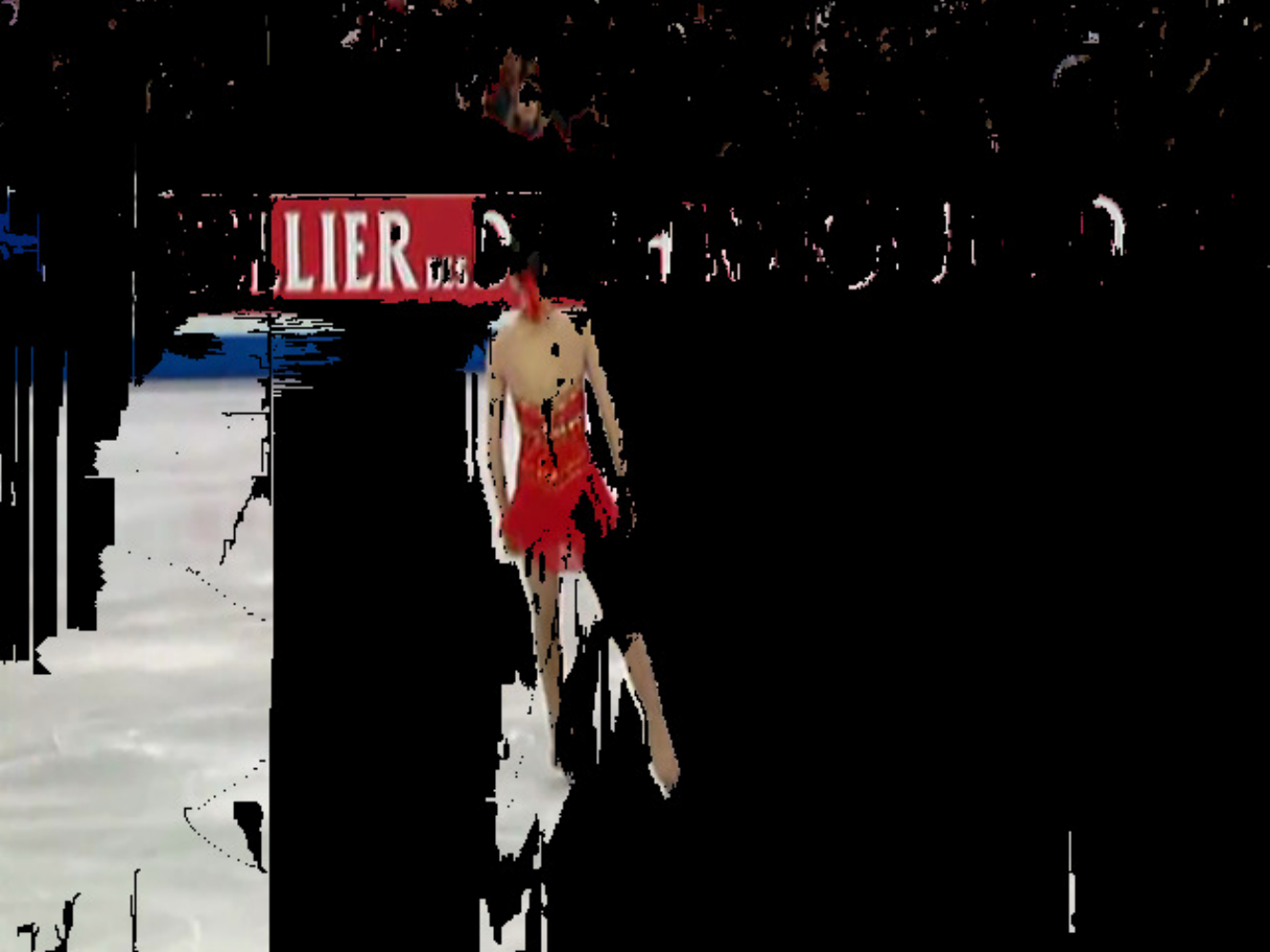}}
        \subfloat{\includegraphics[width=0.15\textwidth]{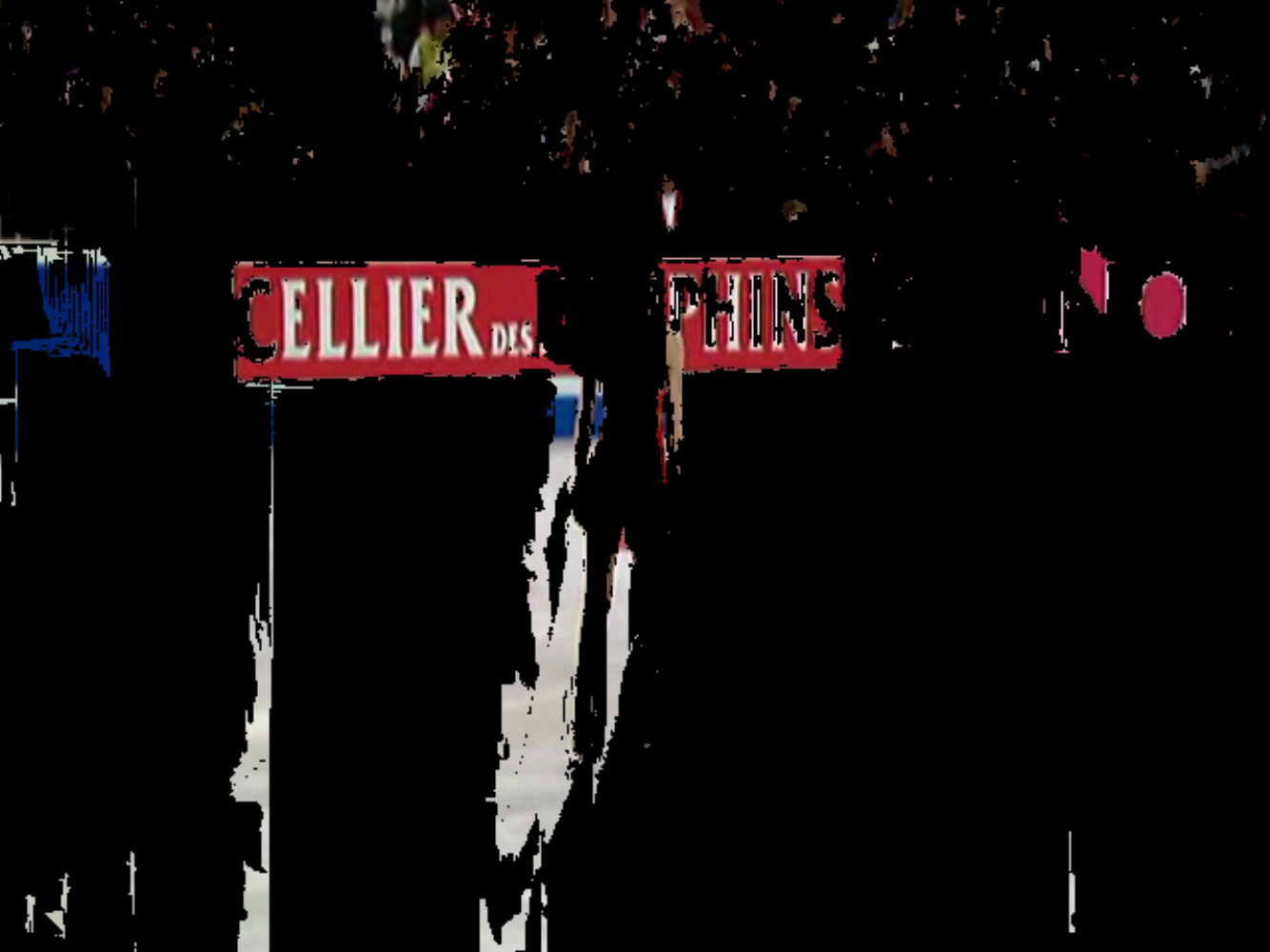}}
        \subfloat{\includegraphics[width=0.15\textwidth]{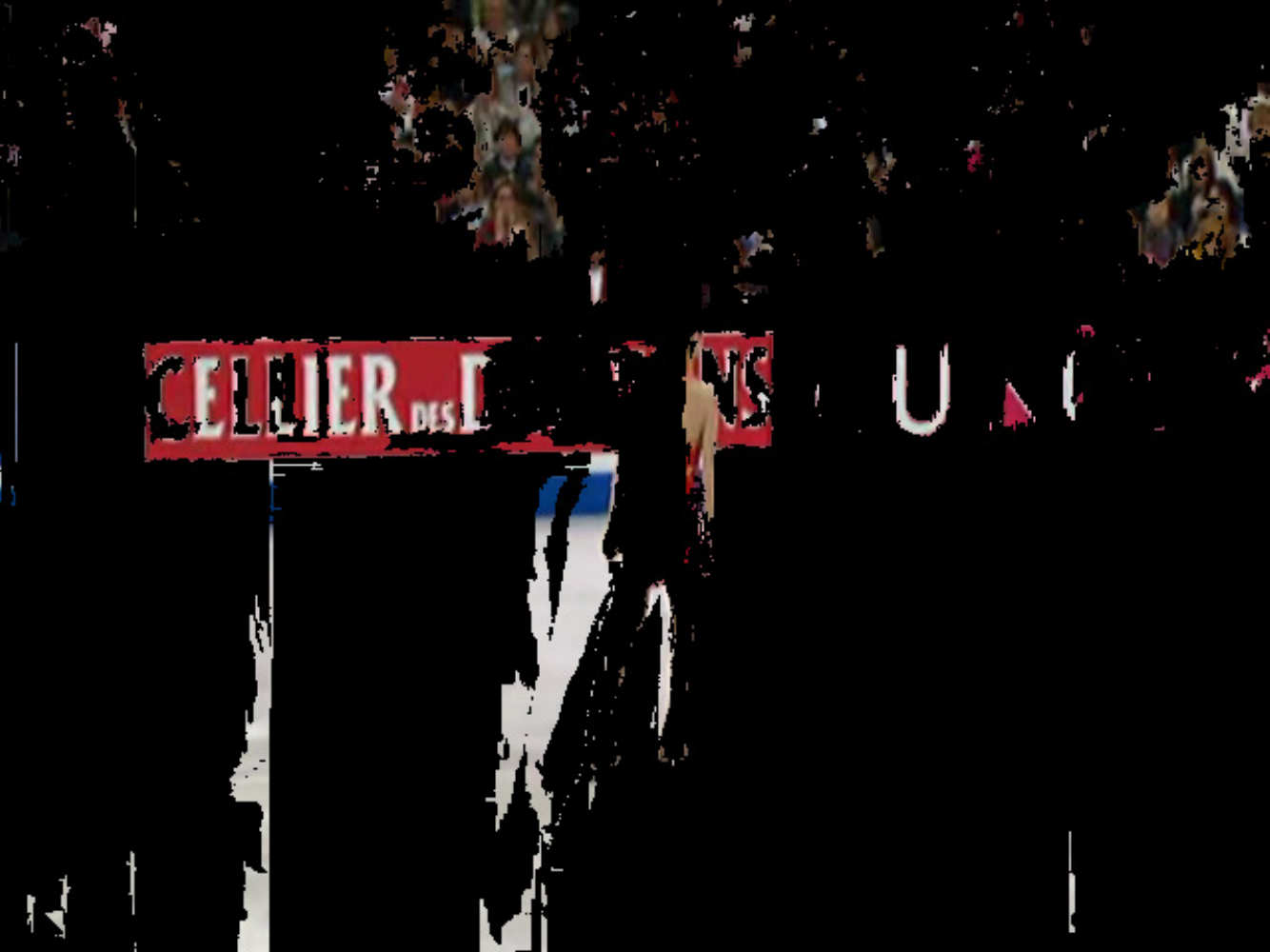}} \\
        \vspace{-3mm} 
        \subfloat{\includegraphics[width=0.15\textwidth]{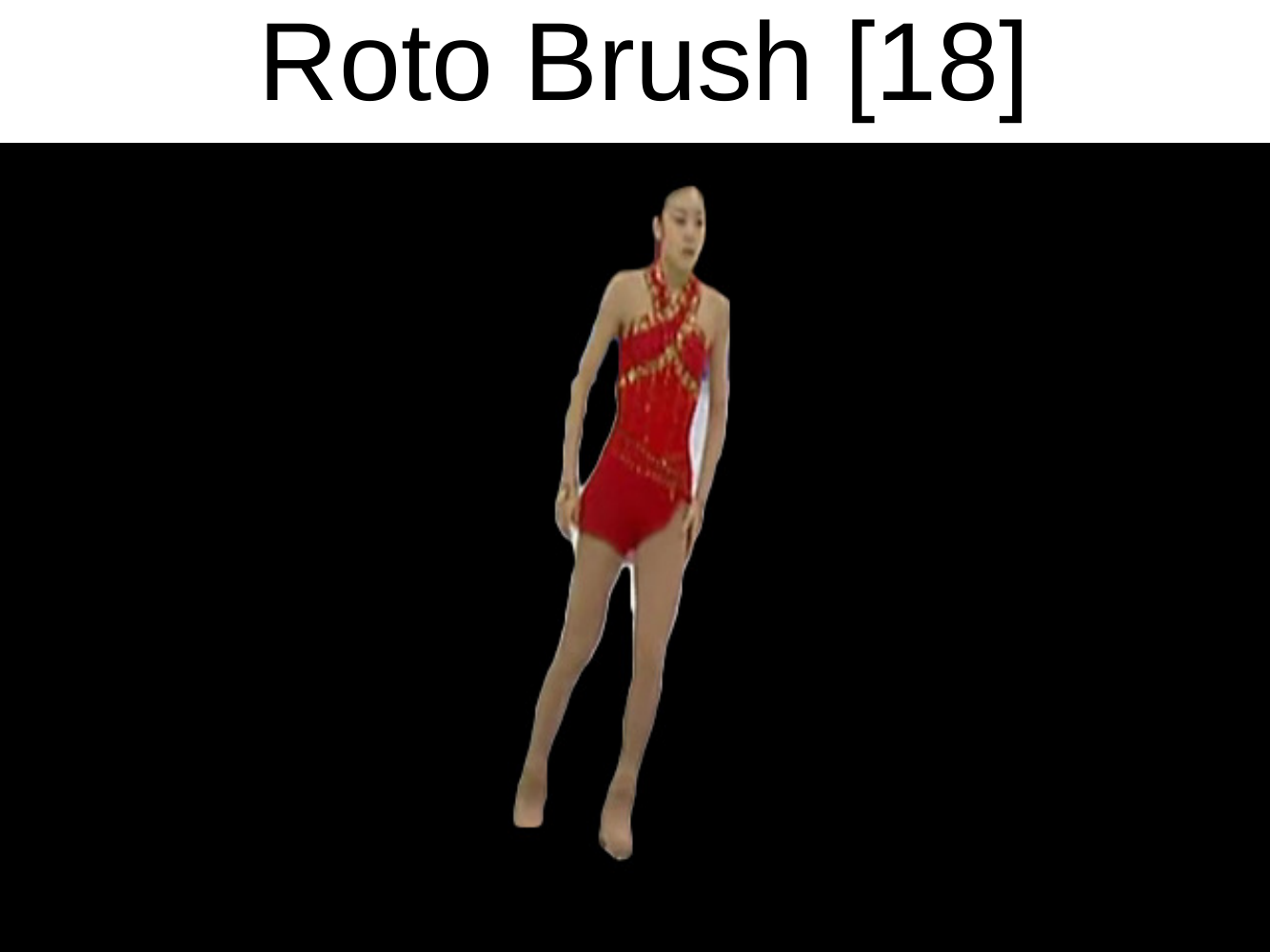}}
        \subfloat{\includegraphics[width=0.15\textwidth]{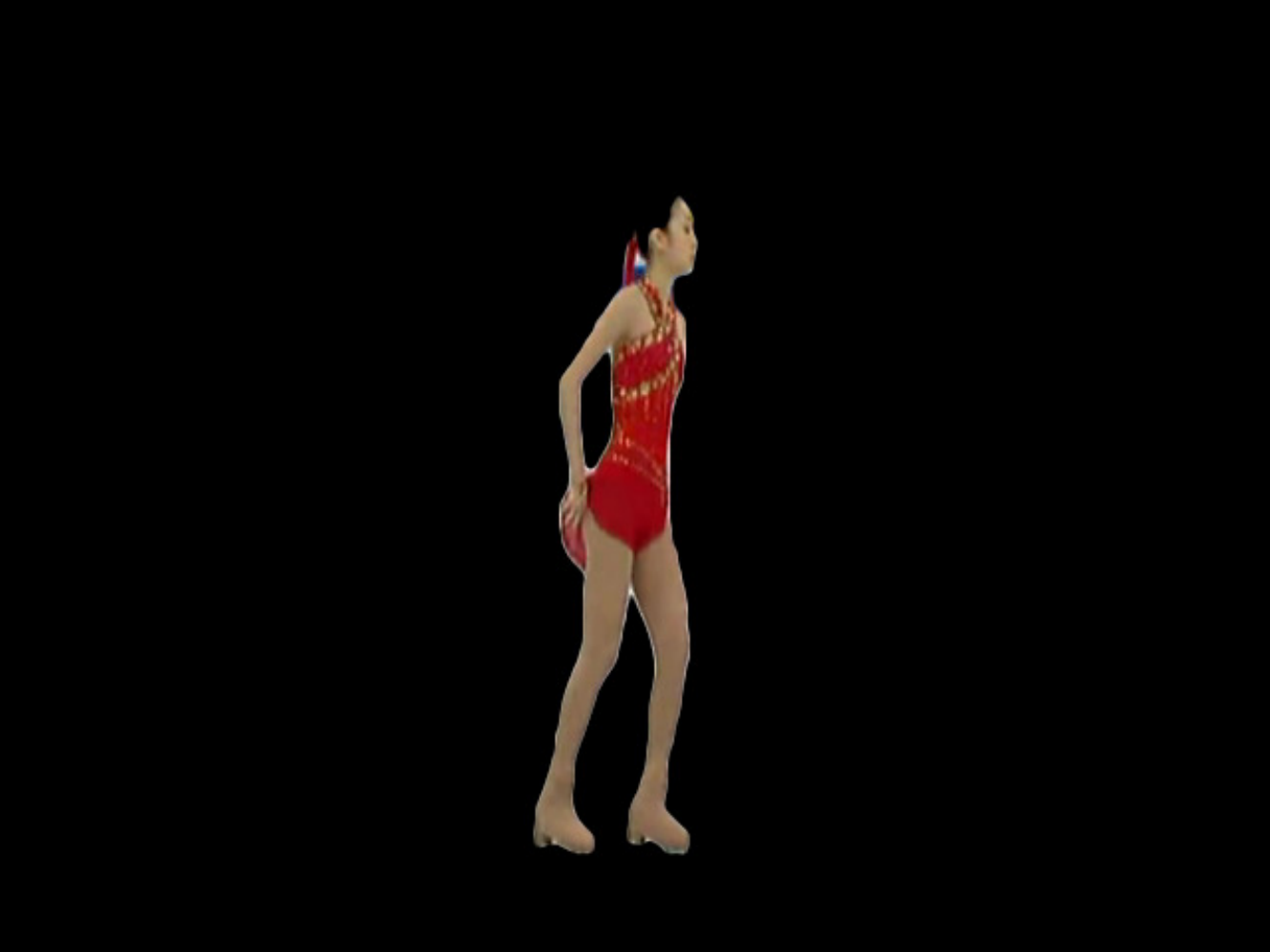}}
        \subfloat{\includegraphics[width=0.15\textwidth]{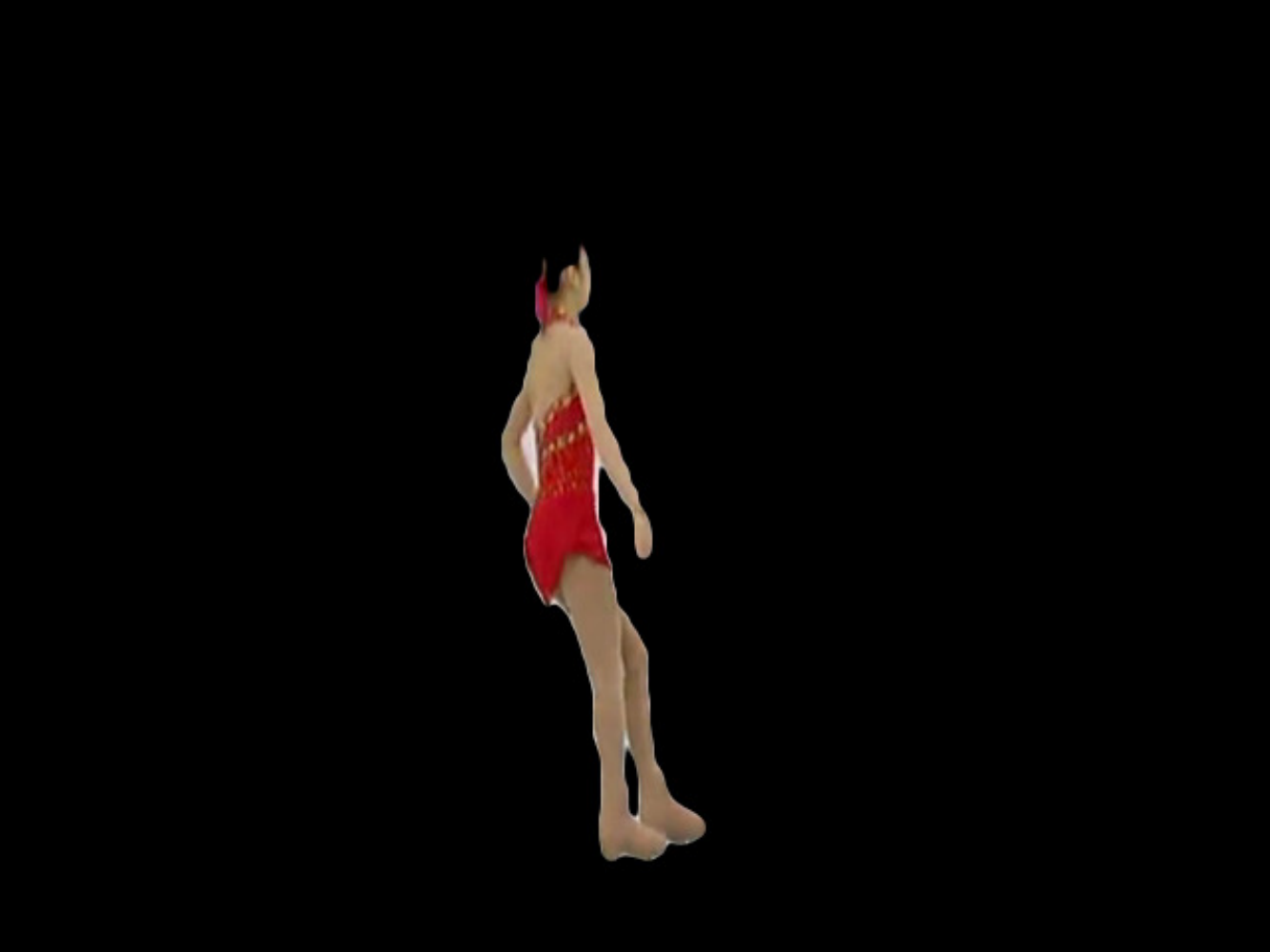}}
        \subfloat{\includegraphics[width=0.15\textwidth]{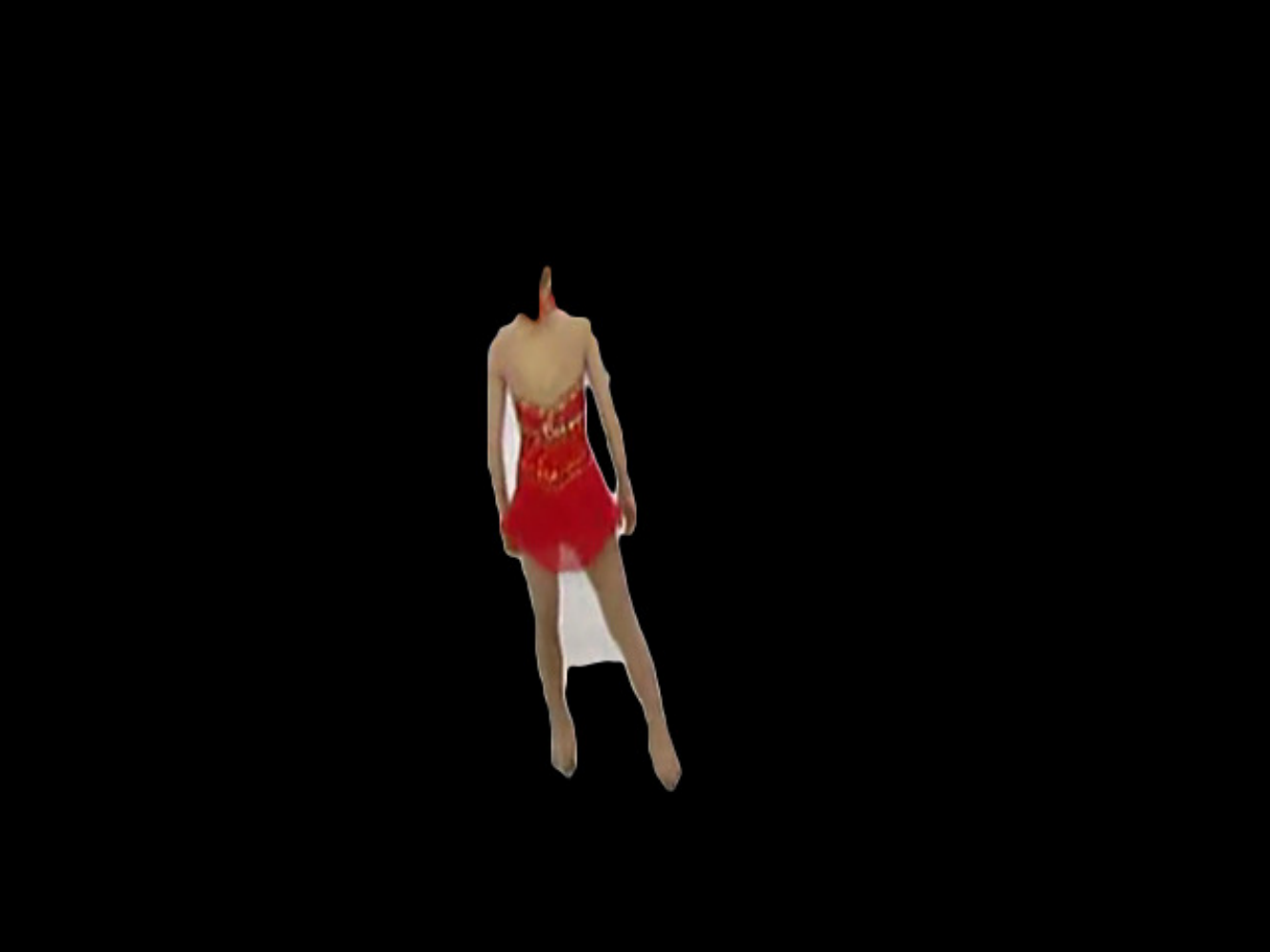}}
        \subfloat{\includegraphics[width=0.15\textwidth]{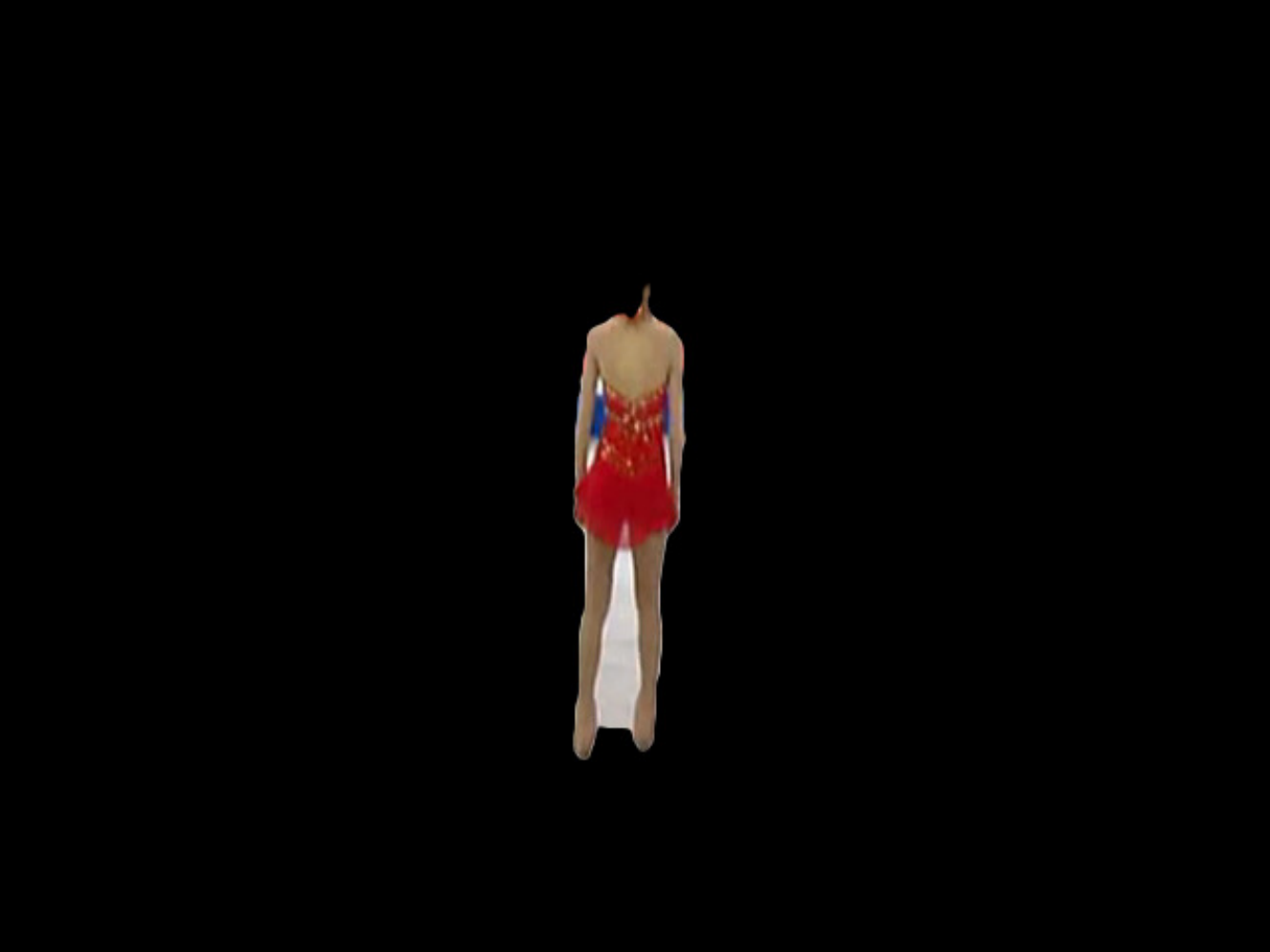}}
        \subfloat{\includegraphics[width=0.15\textwidth]{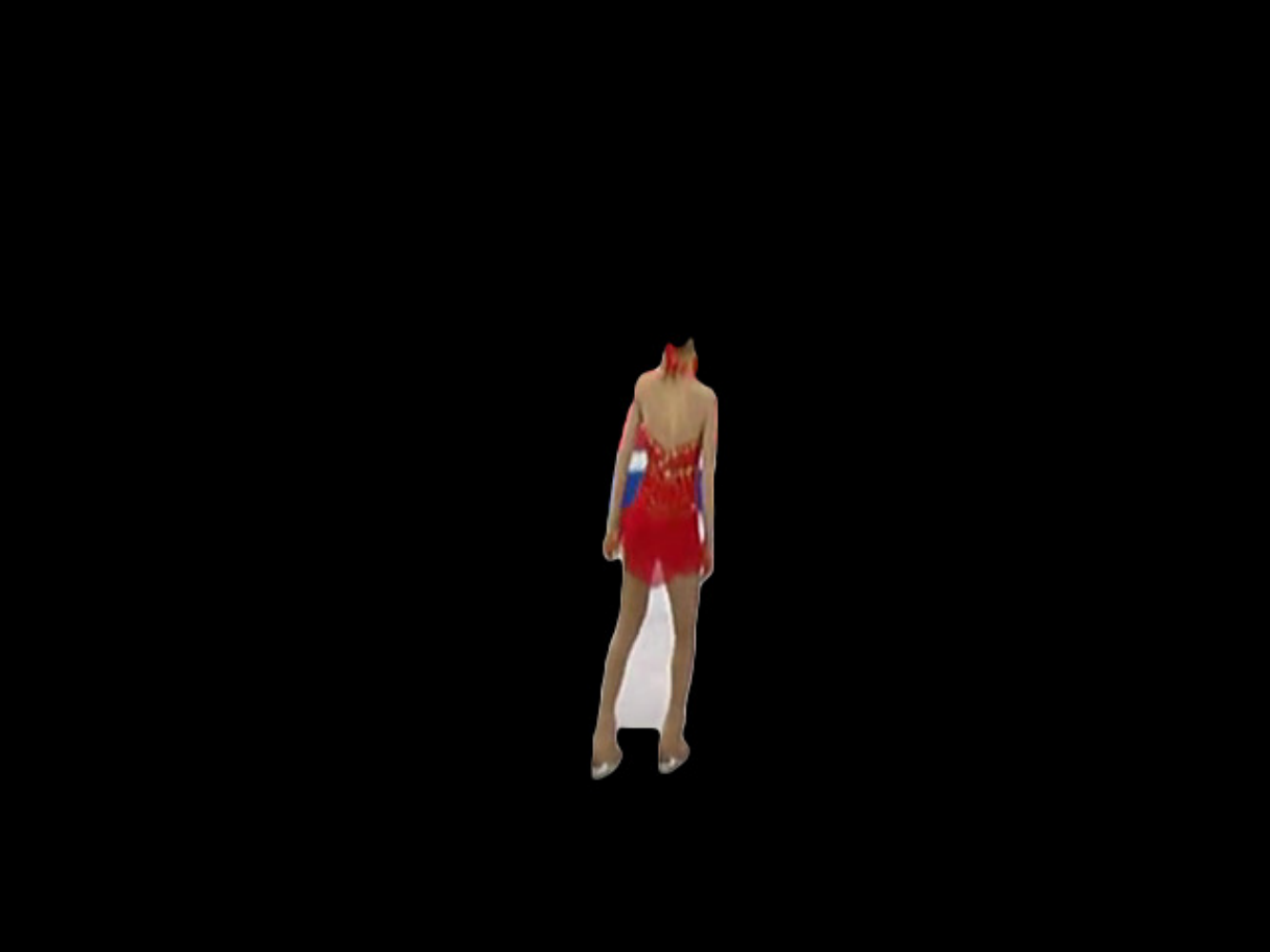}} \\
 \vspace{-3mm}
        \subfloat[Frame 16]{\includegraphics[width=0.15\textwidth]{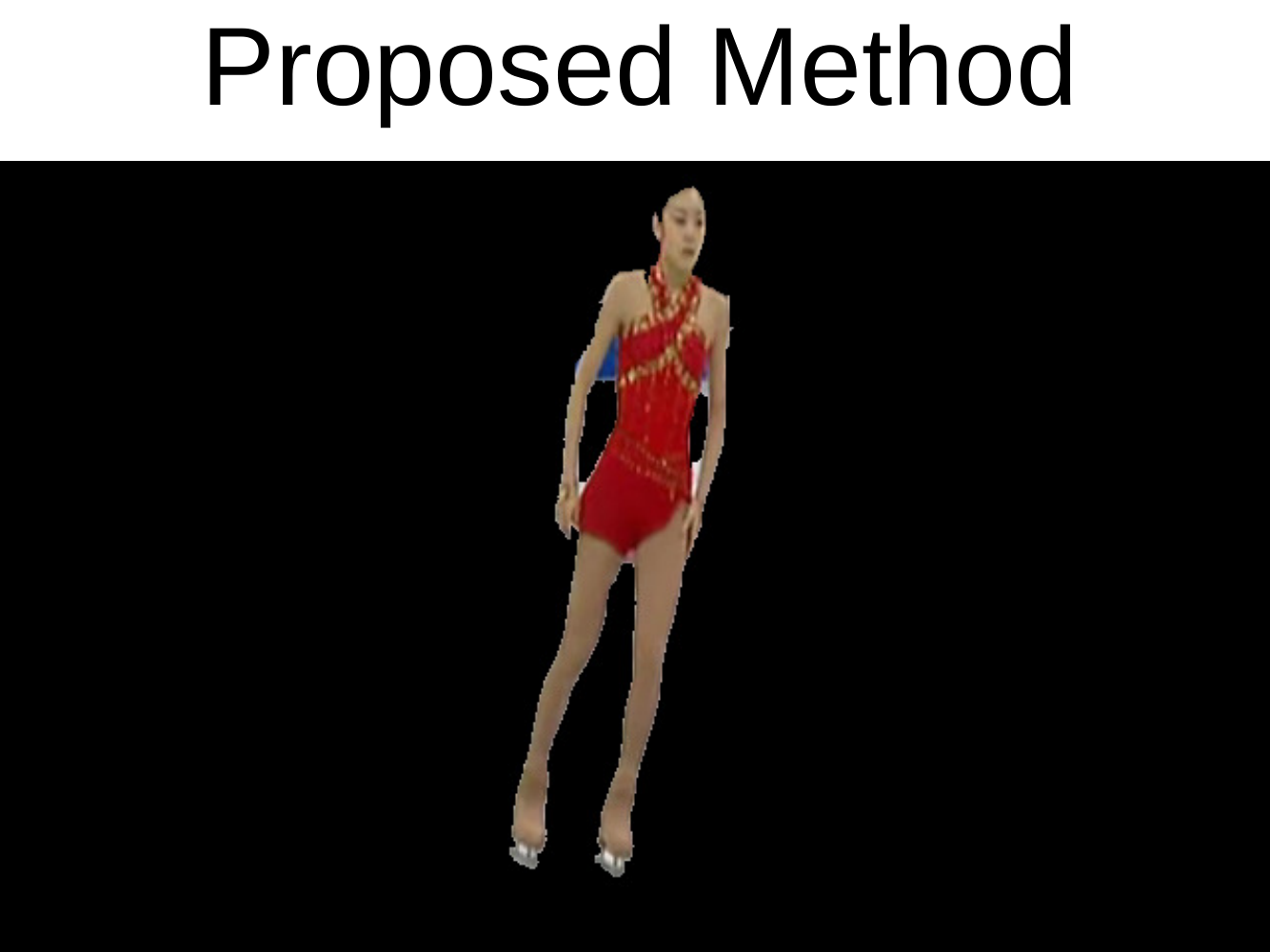}}
        \subfloat[Frame 31]{\includegraphics[width=0.15\textwidth]{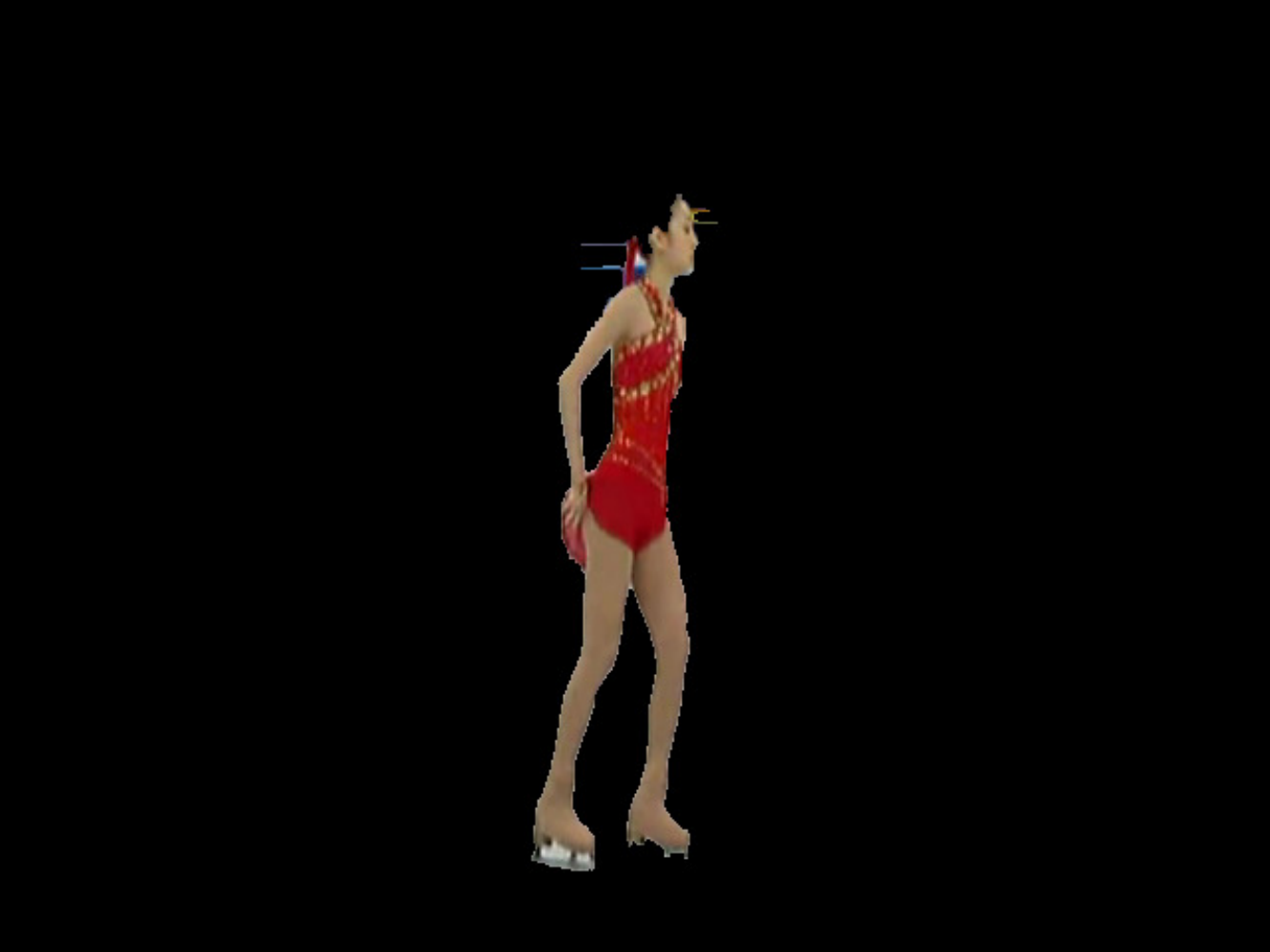}}
        \subfloat[Frame 46]{\includegraphics[width=0.15\textwidth]{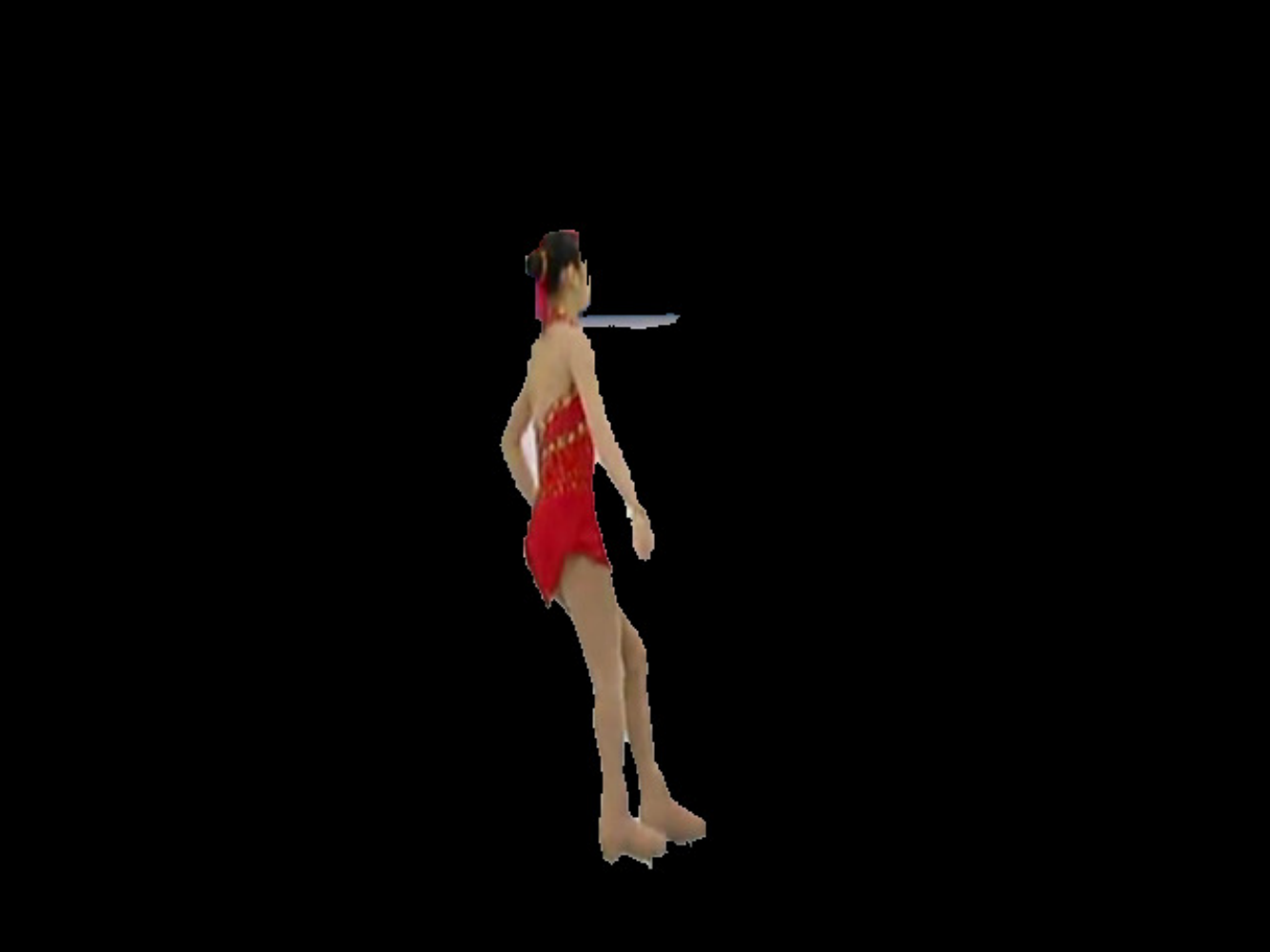}}
        \subfloat[Frame 61]{\includegraphics[width=0.15\textwidth]{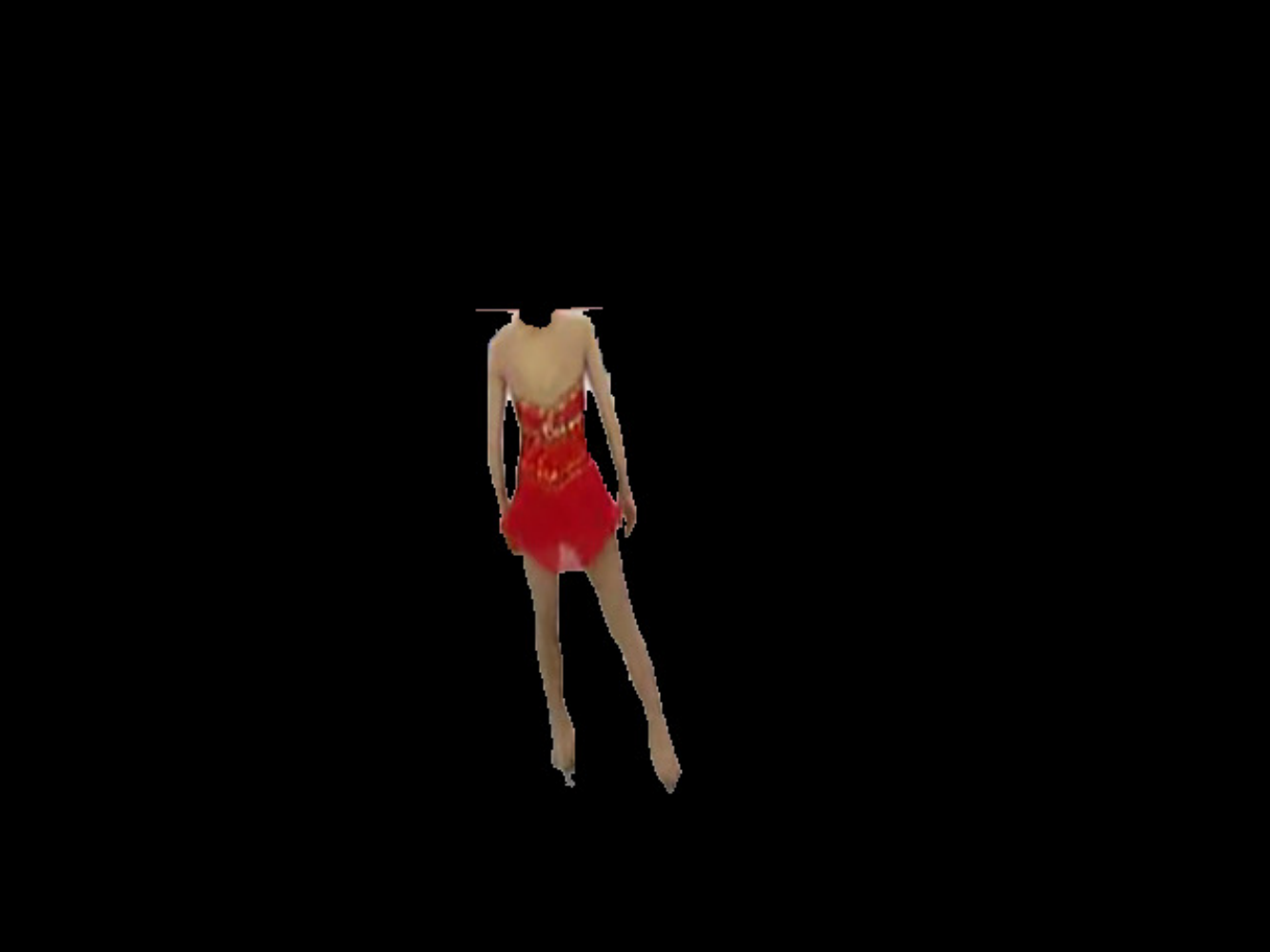}}
        \subfloat[Frame 76]{\includegraphics[width=0.15\textwidth]{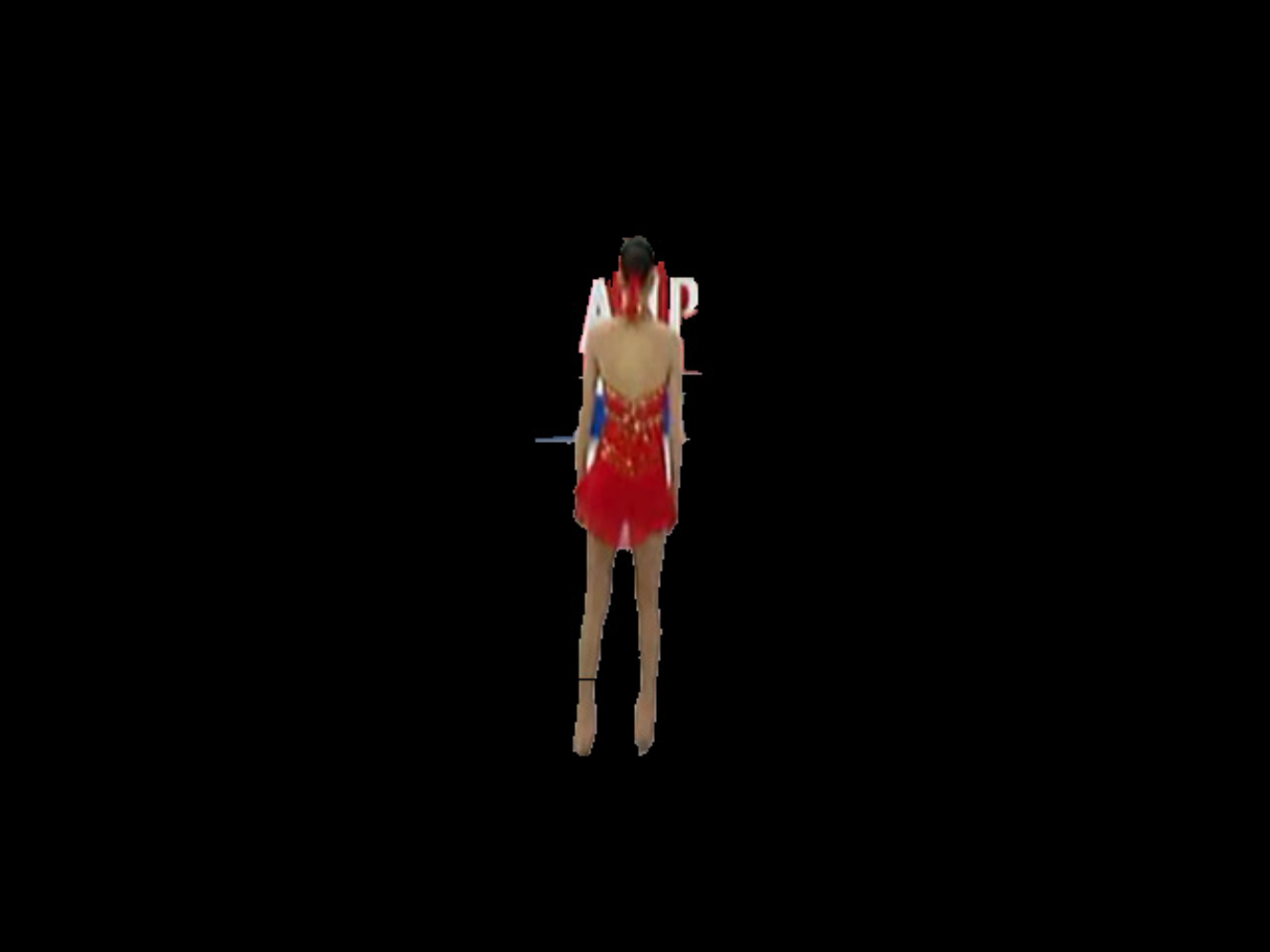}}
        \subfloat[Frame 91]{\includegraphics[width=0.15\textwidth]{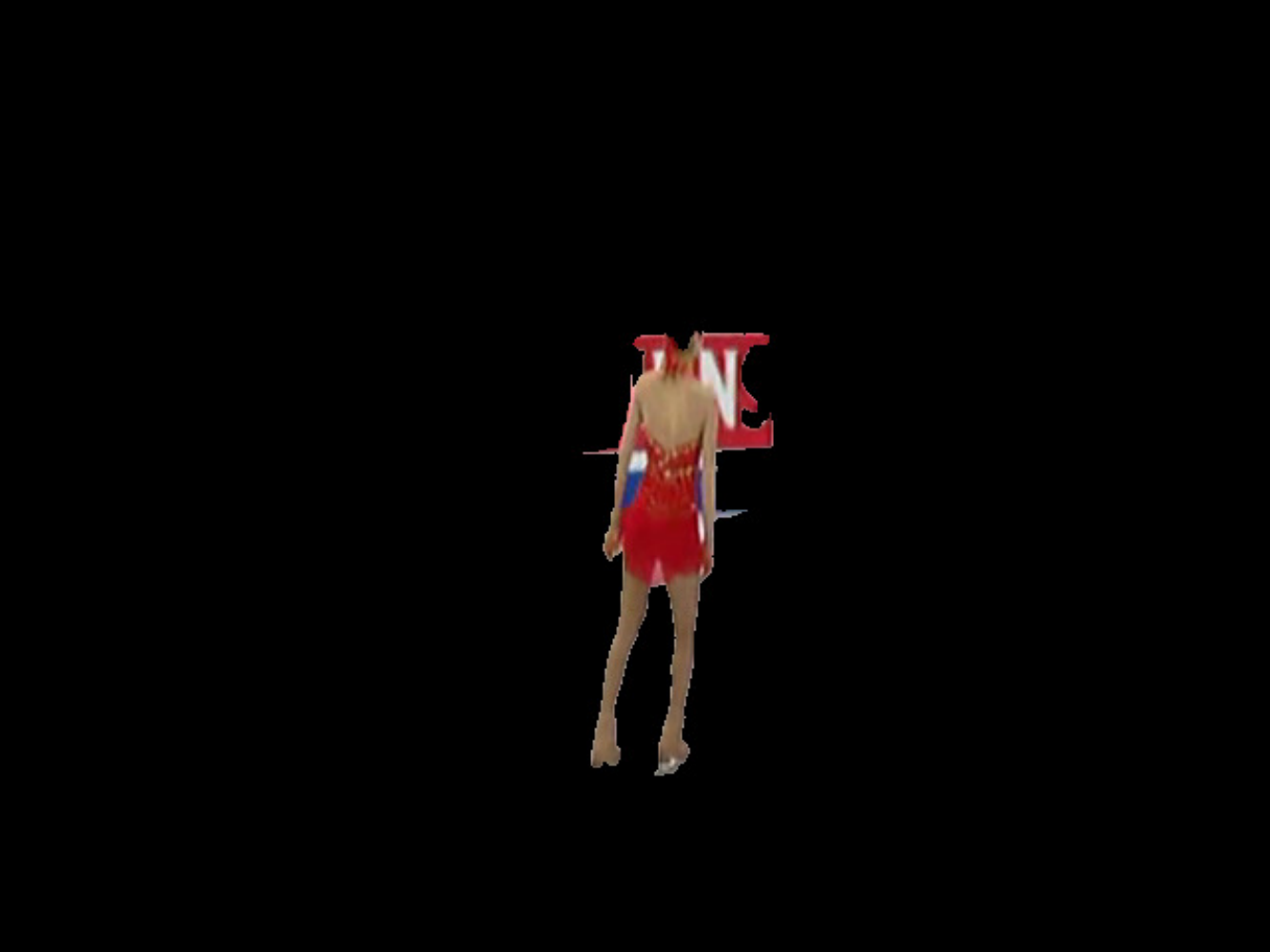}}
\vspace{-1mm}
        \caption{Visual comparison of interactive video segmentation algorithms for \emph{iceSkater} sequence. Top row shows the original video, second row shows the result of \emph{ geodesic video segmentation} tool \cite{geo2}, third row shows the result of \emph{roto brush} tool \cite{roto}, and bottom row shows the result of proposed algorithm. Superior performance of the \emph{roto brush} and proposed method against \emph{ geodesic video segmentation} is clear in all frames. Superior performance of the proposed method against \emph {roto brush} can also be observed in Frame 61 and 76.} \label{qual}
\vspace{-4mm}
\end{figure*}

For the subjective comparison of the algorithms, we have used the dataset used in \cite{graphVideo}. For the \emph{iceSkater} sequence, the resultant segmentation and input video is presented in Figure \ref{qual}. For the subjective comparison, only interactive video object segmentation methods are included. 
\captionsetup[subfloat]{labelformat = empty}
\captionsetup[subfloat]{labelformat = parens	}
It can be observed in Figure \ref{qual}, \emph{ roto brush tool} \cite{roto} and the proposed method have similar performances. On the other hand, geodesic video matting tool \cite{geo2} fails to propagate the segmentation results. The main reason for this performance drawback is its single pixel support region. The motion information is also discarded in the geodesic video matting tool. For the first 50 frames, both the proposed method and \emph{roto brush} \cite{roto} gives near optimal segmentation results. Then, both algorithms segments part of the background as foreground erroneously. In addition to these, at frames 76 and 91, the region around legs of the ice skater is segmented as foreground by \emph{roto brush}. The main reason for this is the shrinking bias of MRF energies \cite{boykovMethod}. However, the proposed method handles shrinking bias successfully due to its wide support region. Hence, one can conclude that the segmentation quality of the proposed method is slightly superior than the \emph{roto brush} \cite{roto}.

\begin{figure*}[t]
        \centering
        \subfloat[Birdfall]{\includegraphics[width=0.3\textwidth]{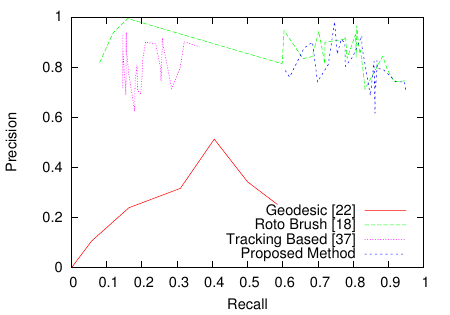}}
        \subfloat[Cheetah]{\includegraphics[width=0.3\textwidth]{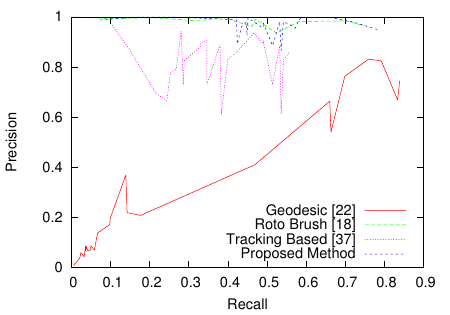}}
        \subfloat[Girl]{\includegraphics[width=0.3\textwidth]{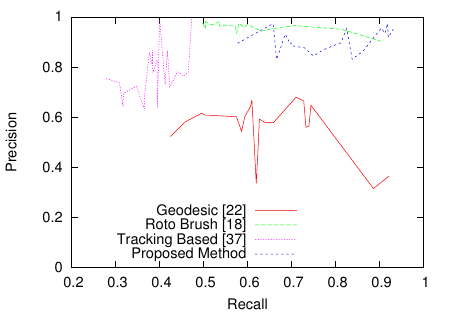}} \\
        \vspace{-3mm} 
		\subfloat[Monkey]{\includegraphics[width=0.3\textwidth]{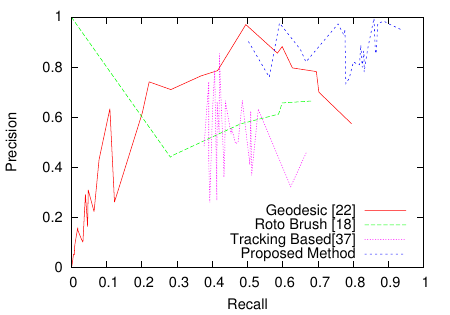}}
        \subfloat[Penguin]{\includegraphics[width=0.3\textwidth]{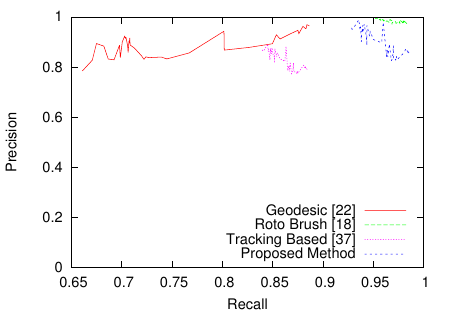}}
        \subfloat[Parachute]{\includegraphics[width=0.3\textwidth]{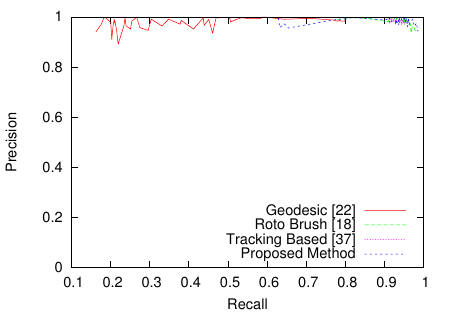}}
        \caption{Precision-Recall curves for SegTrack\cite{segtrack} dataset (A curve near to top-right corner indicates better performance). Precision-recall curves suggest that for interactive video segmentation problem, the proposed method shows either superior or comparable performances against the available methods in the literature.} \label{pr}
\end{figure*}

For the quantitative comparison of the algorithms, we have used \emph {Segtrack} dataset \cite{segtrack}. The initial frame is segmented by using the same user interaction, and the algorithms are executed for the rest of the video. Then, pixel-based precision and recall values are computed for each frame  via the following relations $precision=\frac{tp}{tp+fn}$ and $recall=\frac{tp}{tp+fp}$; where $tp$ is true positive, $fp$ is false positive and $fn$ is false negative. The resulting precision recall curves for each video sequence is plotted in Figure \ref{pr}. Due to the space constraints, visual comparisons are not included in the paper. However, they can be reached through the web-page of the paper (http://www.ozansener.net/mrf-propagation/).

\begin{figure*}[t]
\vspace{-4mm}
        \centering
        \subfloat[Girl Frame 16]{\includegraphics[width=0.3\textwidth]{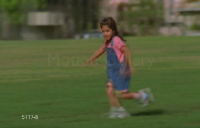}}
        \subfloat[Monkey Frame 19]{\includegraphics[width=0.3\textwidth]{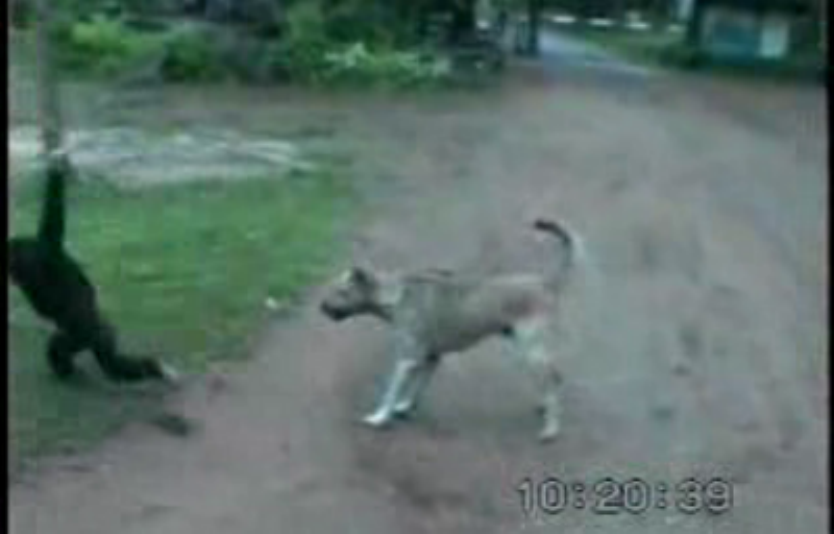}}
        \subfloat[Penguin Frame 1]{\includegraphics[width=0.3\textwidth]{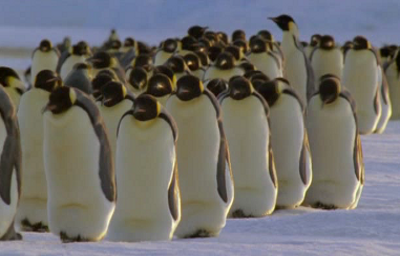}}
        \caption{Visualization of the motion blur and color characteristic in SegTrack \cite{segtrack} dataset. Motion blur in the \emph{ Girl} sequence is clearly visible in (a). The black monkey in \emph {Monkey} sequence, is at the image boundary; therefore, local windows have no color information. Moreover; in \emph{ penguin} dataset, ground truth is selected as a single penguin. Hence, interactive image segmentation algorithm fails due to the repetitive structure of the image.}
        \vspace{-5mm} \label{blur}
\end{figure*}

It can be observed from Figure \ref{pr} that in all videos, both the proposed method and \emph{ roto brush} \cite{roto} have superior performance against \emph{ geodesic method} \cite{geo2}. The main reason for this result is the incorporation of motion information in \cite{roto} and implicit usage of spatial information and wide support region (whole frame) for the proposed method. This result clearly indicates the necessity of the spatial information usage in video segmentation problem. On the other hand, \emph{tracking based method} have competitive precision values and slightly worse recall values. Moreover, proposed method over performs \emph{tracking based method} in all videos.

For \emph{BirdFall} sequence, the proposed algorithms has better recall values then \emph{ roto brush} \cite{roto}. For the points having recall value higher than 0.6, the algorithms show almost same performance. For the \emph{Cheetah} sequence, \emph{ roto brush} and the proposed method have again similar performances. For the \emph{Girl} sequence, \emph{ roto brush} \cite{roto} outperforms the proposed method. This behaviour is due to the high motion blur in the sequence. For the motion blur case, color differences are low even at the object boundaries; therefore, the filter coefficients are not computed properly. An example of the motion blur in the dataset is shown in Figure \ref{blur}. Hence, in \emph{Girl} sequence, there is almost no edge information to be used in the frames due to the motion blur; therefore, the proposed method fails. 

For \emph{Monkey} sequence, the proposed method out performs \emph{ roto brush} \cite{roto}. The motion occurs around the frame boundary; therefore, there is not enough local color information to be used for \emph{ roto brush} \cite{roto}. This characteristic of the video is also visualized in Figure \ref{blur}. For \emph{Penguin} sequence, utilized interactive image segmentation algorithm fails to segment the initial frame. Hence, the performance drawback is not due to the proposed energy propagation method. An example image is also shown in Figure \ref{blur}. Finally, for the \emph{Parachute} sequence, both \emph{ roto brush} \cite{roto} and the proposed method have precision and recall values higher than 0.9; both algorithms have similar precision values; however, the proposed method have better recall values.
\subsubsection{Computation Time}
In order to compare the computational efficiencies of the algorithms, we have used the \emph{ SegTrack} \cite{segtrack} dataset. Each sequence in the dataset is rescaled to 640x480 resolution, and, segmentation time for each frame is computed for all sequences. The mean computation time for each frame is summarized in Table 1.
\begin{table}
\centering
\caption{Overall Computation Time per Frame}
  \begin{tabular}{ |c | c | c| }
    \hline
       Geodesic Segmentation \cite{geo2} & Roto Brush \cite{roto} & Proposed Method\\ \hline
    2,7 sec & 2,3 sec & 1,3 sec\\
    \hline
  \end{tabular}
  \vspace{-4mm}
\end{table}

As it can be observed from the Table 1, the proposed method has the highest computational efficiency. It should also be noted that, the significant part of the computation time for the proposed method is consumed by SLIC \cite{slic} algorithm. Convergence of SLIC \cite{slic} algorithm takes approximately 0.9 second per frame. Hence, by the introduction of efficient over-segmentation algorithms, it is possible to reach much better computation time for the proposed method.
 
\begin{figure}[h!]
\vspace{-5mm}
        \centering
        \subfloat[Recall vs Computation Time]{\includegraphics[width=0.23\textwidth]{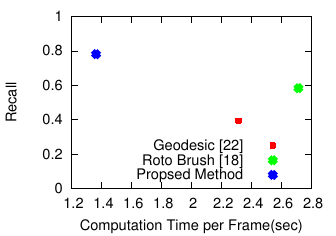}} 
		\subfloat[Precision vs Computation Time]{\includegraphics[width=0.23\textwidth]{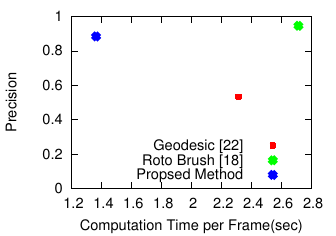}} 
		\caption{Recall and Precision versus Computation Time. The results suggest that the proposed method reaches twice the efficiency with superior recall and comparable precision values.}
		\vspace{-2mm}
		\label{prct}
\end{figure}

In order to visualize the performance vs time trade-off for all the algorithms, we have also performed performance vs time comparisons. For the \emph{ SegTrack} \cite{segtrack} dataset, we have computed mean precision and mean recall values as well as the corresponding computation times. Precision vs computation time and recall vs computation times are presented in Figure \ref{prct}.
As it can be observed from Figure \ref{prct}.a, the proposed method has better recall and computation time, when it is compared against other algorithms. In Figure \ref{prct}.b, \emph{ roto brush} \cite{roto} has slightly better precision values with a higher computational burden. One can conclude that the proposed method reaches segmentation quality of \emph{ roto brush} \cite{roto} with much higher computational efficiency. On the other hand, \emph{tracking based method} \cite{segtrack} is not included in this plots since its computation time per frame is around 1 minute. Moreover, computation time of the \emph{tracking based method} is not comparable with other algorithms. Indeed, its computational complexity is not affordable in any interactive system. 
\subsubsection{Number of Interacted Frames}
In order to analyse the effect of the number of interacted frames, we designed another experiment. Since the proposed algorithm is designed for an interaction in the initial frame, only the first frame is used for the interaction for the proposed method. Whereas; for the experiments on \emph{roto brush} \cite{roto} and \emph{geodesic method} \cite{geo2}, we let user interact on as many frames as he/she wants. User initially interacts with the first frame and see the segmentation result for the rest of the frames; then, user updates the segmentation by choosing another key-frame and giving interaction for it and so on. Within the experiments, SegTrack \cite{segtrack} data set is used. For each video, precision, recall as well as number of interacted frames is recorded.

\begin{figure}[h!]
\vspace{-3mm}
        \centering
	\includegraphics[width=0.44\textwidth]{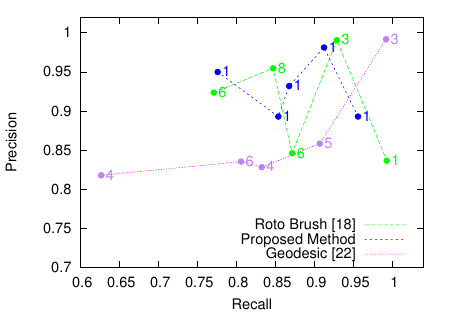}
		\caption{Number of interacted frames, precision and recall values for each sequence in SegTrack \cite{segtrack} dataset. Each point represents a video sequence; precision and recall values are y and x coordinates respectively and the number of interacted frames are shown right next to the data points. First 30 frames of each sequence is used for the experiment; hence, number of frames in each sequence is equal.}
		\vspace{-3mm}
		\label{ptt}
\end{figure}
In Figure \ref{ptt}, precision-recall curve over SegTrack \cite{segtrack} dataset is shown. Precision-recall curve is augmented with the number of interacted frames for each sequence. As shown in the Figure \ref{ptt}, proposed method and \emph{roto-brush} \cite{roto} have similar precision and recall values. On the other hand, \emph{roto-brush} requires interaction on 3 to 8 frames for comparable performance with the proposed method which only requires interaction on the first frame. Moreover, \emph{geodesic method} \cite{geo2} can not reach to the performance level of the proposed method and \emph{roto-brush} even with high number of interactions.

\subsection{Dynamic Bilateral Graph-Cut}
In order to experiment the computational improvements obtained by dynamic computation of graph-cut in filtering scenario, we have performed MRF energy propagation via both conventional min-cut/max-flow method \cite{boykovMethod} and the proposed dynamic-graph cut. As explained in Section 4, dynamic graph-cut presented in \cite{dynGCut} is not applicable to our scenario, since the structure of the graph in the proposed method changes significantly due to the over-segmentation. 

\begin{figure}[h!]
        \centering
        \includegraphics[width=0.44\textwidth]{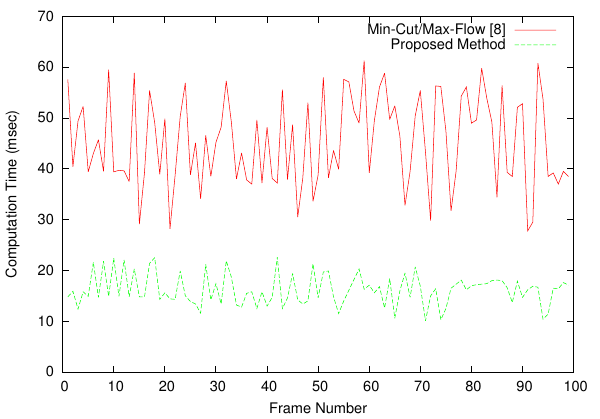}
		\caption{Computation time for dynamic graph-cut of each frame. The proposed improvements results in around 3 times efficiency increase.}
		\vspace{-3mm}
		\label{dyn}
\end{figure}

For the min-cut/max-flow, we propagate MRF energy and applied min-cut/max-flow \cite{boykovMethod}. For the dynamic graph-cut, we use the proposed dynamic method. Only the graph-cut execution times for each frame is  plotted. As shown in Figure \ref{dyn}, the proposed improvements results in a significant decrease in the computation time.

\subsection{Automatic Video Segmentation Speed-up}
As explained in Section 1 and 2, the proposed energy propagation tool can actually be used in order to speed-up any automatic video segmentation tool. We have experimented this scenario on a recently proposed automatic segmentation algorithm Key-Segments \cite{keysegments}. In Key-Segments algorithm, graph-based 2D clustering of frames are performed to generate as many hypothesis as possible. Then, these hypothesizes are ranked by using a saliency-like measure and this measure is computed by using both shape, texture and motion information. Both extraction of optical flow vectors and computation of saliency-like measure is computationally expensive. In our setup, Key-Segments \cite{keysegments} method is applied only to initial $k$ frames ($k$=3). Then, the color information of both foreground and background is obtained via EM procedure in terms of Gaussian Mixture Models (GMM). Then, the energy function is defined as in the case of interactive image segmentation \cite{mmpd} only for initial $k$ frames. No infinite terminal weight is used due to the lack of definite user interaction. After the definition of the energy function for the first few frames, the proposed energy propagation and minimization tool is used to propagate and solve the segmentation problem for the rest of the video.

\begin{figure}[th]
        \centering
        \subfloat[Birdfall]{\includegraphics[width=0.23\textwidth]{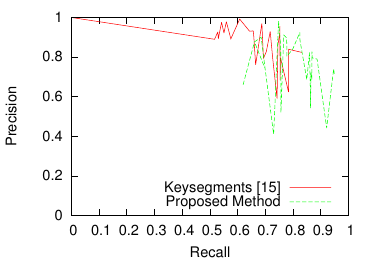}}
        \subfloat[Cheetah]{\includegraphics[width=0.23\textwidth]{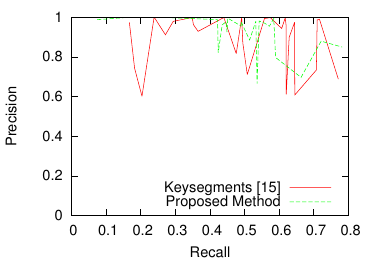}} \\ \vspace{-4mm}
        \subfloat[Girl]{\includegraphics[width=0.23\textwidth]{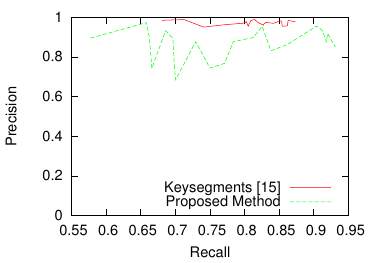}} 
		\subfloat[Monkey]{\includegraphics[width=0.23\textwidth]{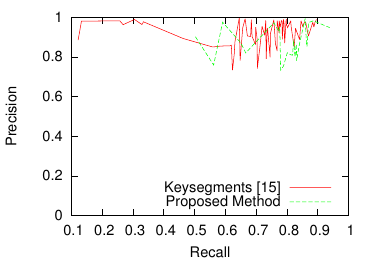}} \\ \vspace{-4mm}
        \subfloat[Parachute]{\includegraphics[width=0.23\textwidth]{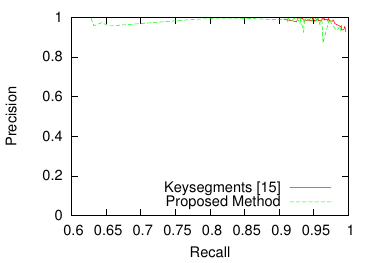}}
        \caption{Precision-recall curves for automatic video segmentation on SegTrack \cite{segtrack} dataset. The proposed method does not cause significant performance drawback in any of the videos, except \emph{Girl} sequence. The reason for this performance drawback is motion blur.} \label{prAuto}
        \vspace{-4mm}
\end{figure}

In order to analyse the performance of these speed-up, we have used \emph{ SegTrack} \cite{segtrack} dataset. As explained in \cite{keysegments}, \emph{Penguin} dataset is discarded due to the ambiguous ground truth. For the initial 3 frames, Key-Segments algorithm is performed and for the rest of the video, the proposed algorithm is used. The computation time and precision-recall values for each frame of each video is recorded and the resulting precision-recall curves are plotted in Figure \ref{prAuto}. The average computation time for each frame is also computed in summarized in Table \ref{autoC}. For a fair comparison,  source code distributed with \cite{keysegments} is used, and the proposed method is reimplemented in MATLAB. 

\begin{table}
\centering
\caption{Computation Time per Frame (in MATLAB)}
  \begin{tabular}{ |c | c| }
    \hline
       Key-Segments \cite{keysegments}  & Speed-up Key-Segments (Proposed Method)\\ \hline
    260.6 sec &  4.0 sec\\
    \hline
  \end{tabular}
  \label{autoC}
  \vspace{-8mm}
\end{table}

When the precision recall curves in Figure 6 is considered; except for the \emph{Girl} sequence, the proposed speed-up method performs comparable against the original Key-Segments \cite{keysegments} method. In \emph{Birdfall} sequence, the proposed method has better recall values than the original Key-Segments \cite{keysegments}. Moreover, in \emph{Cheetah} and \emph{Monkey} sequences, both algorithms perform very similar to each other. In \emph{Parachute} sequence, both algorithms performs near optimal and most of the performance drawback of proposed method is due to the superpixel errors caused by SLIC \cite{slic} algorithm. Finally, in \emph{ Girl} sequence the proposed algorithms perform much worse than the original Key-Segments \cite{keysegments}. The main reason of this performance drawback is due to the high amount of motion blur in the sequence. The edge information is used extensively by permeability filter; therefore, lack of edges results in serious performance drawback. Motion blur in the \emph{Girl} sequence is visualized in Figure 6, and discussed in Section V.A.I In conclusion, except for the case of very high motion blur, the proposed speed-up results in similar segmentation quality, when compared to the original segmentation algorithm.

On the other hand, when computational complexity is considered, the proposed algorithm results in around 65 times speed-up when compared with the original algorithm. It is reasonable because main bottleneck of the Key-Segments \cite{keysegments} algorithm is its initial object proposal step and it is avoided in the proposed method. 

\section{Conclusion}
In this paper, we address the problem of video segmentation from the perspective of computational efficiency. We redefine the problem as propagation of MRF energies through frames. We propose a novel technique via bilateral filters in order to increase computational efficiency. We further increase the efficiency by a novel dynamic graph-cut formulation applicable to a linear filtering scenario. The proposed method can either be used as an interactive video segmentation tool or a speed-up tool for automatic video segmentation. The experimental results suggest that in both scenarios, the proposed methods produce segmentations having higher or comparable quality when compared against the leading algorithms from the literature. From the perspective of efficiency, the proposed method results in 2 times increase in the computational efficiency for interactive video segmentation and 65 times increase in the computational efficiency for automatic video segmentation with almost no significant decrease in segmentation quality.

\ifCLASSOPTIONcaptionsoff
  \newpage
\fi
\vspace{-3mm}
\appendices
\section{Proof of the Proposition 1}
\begin{proposition2}
Binary labels obtained by minimizing the MRF energy, resulted after applying bilateral filter on the energy function which is defined via residual graph, is equivalent to minimizing the MRF energy obtained via applying bilateral filter on the original energy function.
\end{proposition2}
\begin{proof}
We prove the proposition by constructing flows at time $t$. The flows are constructed by applying bilateral filter to the flows at time $t-1$. We show that these flows converts the propagated energy function to a propagated residual energy. 

We will consider the flows in terminal and non-terminal weights separately. Furthermore, our proof will be based on edge flows; not source-sink flows. Source-sink flows can be obtained via concatenation of edge flows. 

For the flows on terminal edges, consider $f^{t-1}_{iT}$ ($T$ is either source or sink) as flows in time $t-1$;
\begin{equation}
f^t_{jT} = \frac{1}{\gamma ^t_i}\sum_{v_i \in V^{t-1}} f^{t-1}_{iT} e^{-dis(z^{t-1}_i,z^t_j)}
\end{equation}
For the non-terminal flows, consider $f^{t-1}_{ij}$ as flows in $t-1$;
\begin{equation}
f^t_{ij} =  \frac{1}{\gamma ^t_{ij}} \sum_{v_k \in V^{t-1}} \sum_{v_l\in N(v_k)} f^{t-1}_{kl} e^{-dis(z^t_i,z^{t-1}_k)} e^{-dis(z^t_j,z^{t-1}_l)} 
\end{equation}
It is shown that applying any flow through graph does not change the solution to minimum cut problem \cite{boykovMethod}. Therefore, the solution after applying these flows does not change. The residual weights are obtained after applying these flows as;
\begin{equation}
\begin{aligned}
r^t_{iT}&= U^t(t,z^t_i) - f^t_{iT} \\
&=\frac{1}{\gamma ^t_i}\sum_{v^{t-1}_j  \in V^{t-1}}U^{t-1}(t, z^{t-1}_j)e^{-dis(z^t_i,z^{t-1}_j)}  \\ &\hspace{15mm}
- \frac{1}{\gamma ^t_i}\sum_{v_j \in V^{t-1}} f^{t-1}_{jT} e^{-dis(z^t_i,z^{t-1}_j)}  \\
&=\frac{1}{\gamma ^t_i}\sum_{v^{t-1}_j \in V^{t-1}} r^{t-1}_{jT} e^{-dis(z^t_i,z^{t-1}_j)}  \\ 
\end{aligned}
\end{equation}
A similar relation ($r^t_{ij}= V^t(z^t_i,z^t_j) - f^t_{ij}
$) can also be obtained for the non-terminal weights as well;
\begin{equation}
\begin{aligned}
r^t_{ij}&=\frac{1}{\gamma ^t_{ij}} \sum_{v_k \in V} \sum_{v_l\in N(v_k)} e^{-d(z^t_i,z^{t-1}_k)} e^{-d(z^t_j,z^{t-1}_l)} V (z^{t-1}_k,z^{t-1}_l) \\
&\hspace{5mm}- \frac{1}{\gamma ^t_{ij}} \sum_{v_k \in V^{t-1}} \sum_{v_l\in N(v_k)} f^{t-1}_{kl} e^{-dis(z^t_i,z^{t-1}_k)} e^{-dis(z^t_j,z^{t-1}_l)} \\
&=\frac{1}{\gamma ^t_{ij}} \sum_{v_k \in V} \sum_{v_l\in N(v_k)} e^{-d(z^t_i,z^{t-1}_k)} e^{-d(z^t_j,z^{t-1}_l)} r^{t-1}_{kl}
\end{aligned}
\end{equation}
This result corresponds to applying bilateral filter to the weights in the residual graph. In summary, we prove that propagating flows used in the min-cut/max-flow and recycling them for the next frame, corresponds to propagating residual graph via bilateral filters.
\end{proof}

\bibliographystyle{ieeetr}

\begin{IEEEbiography}[{\vspace{-7mm}\includegraphics[width=1in,height=1.25in,clip,keepaspectratio]{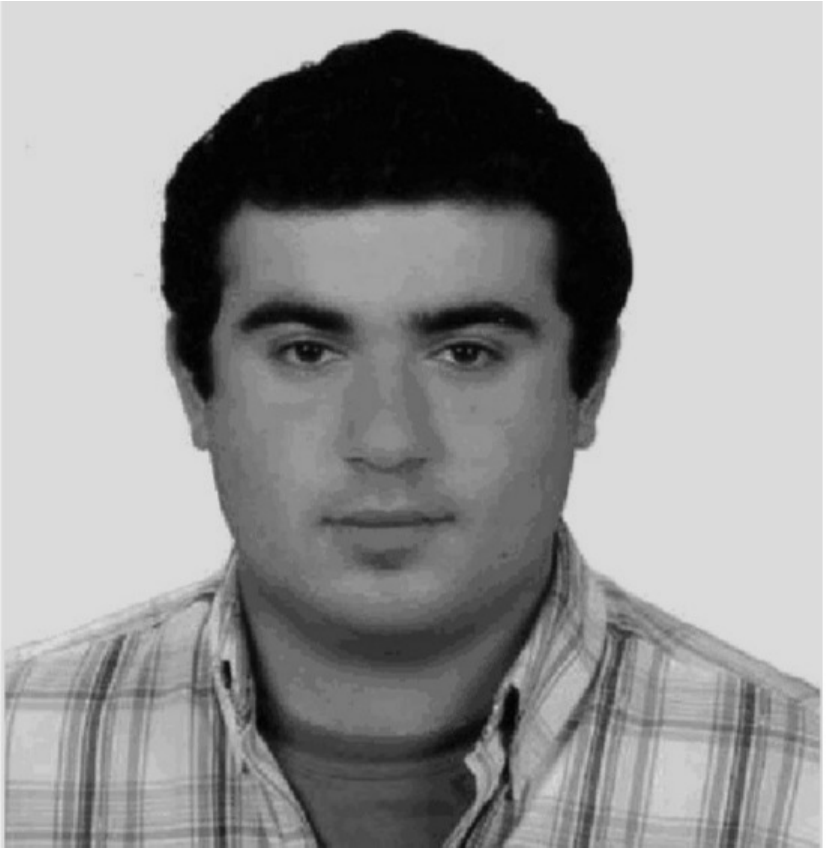}}]{Ozan Sener}
received the B.Sc. and M.Sc. degree from the Electrical Engineering Department of Middle East Technical University, Ankara, Turkey in 2010 and 2013, and is currently pursuing the Ph.D. degree at Electrical and Computer Engineering Department of Cornell University. 

His research interests are graph based methods for multimedia, machine learning and robotics. 
\end{IEEEbiography}
\vspace{-5mm}
\begin{IEEEbiography}[{\includegraphics[width=1in,height=1.25in,clip,keepaspectratio]{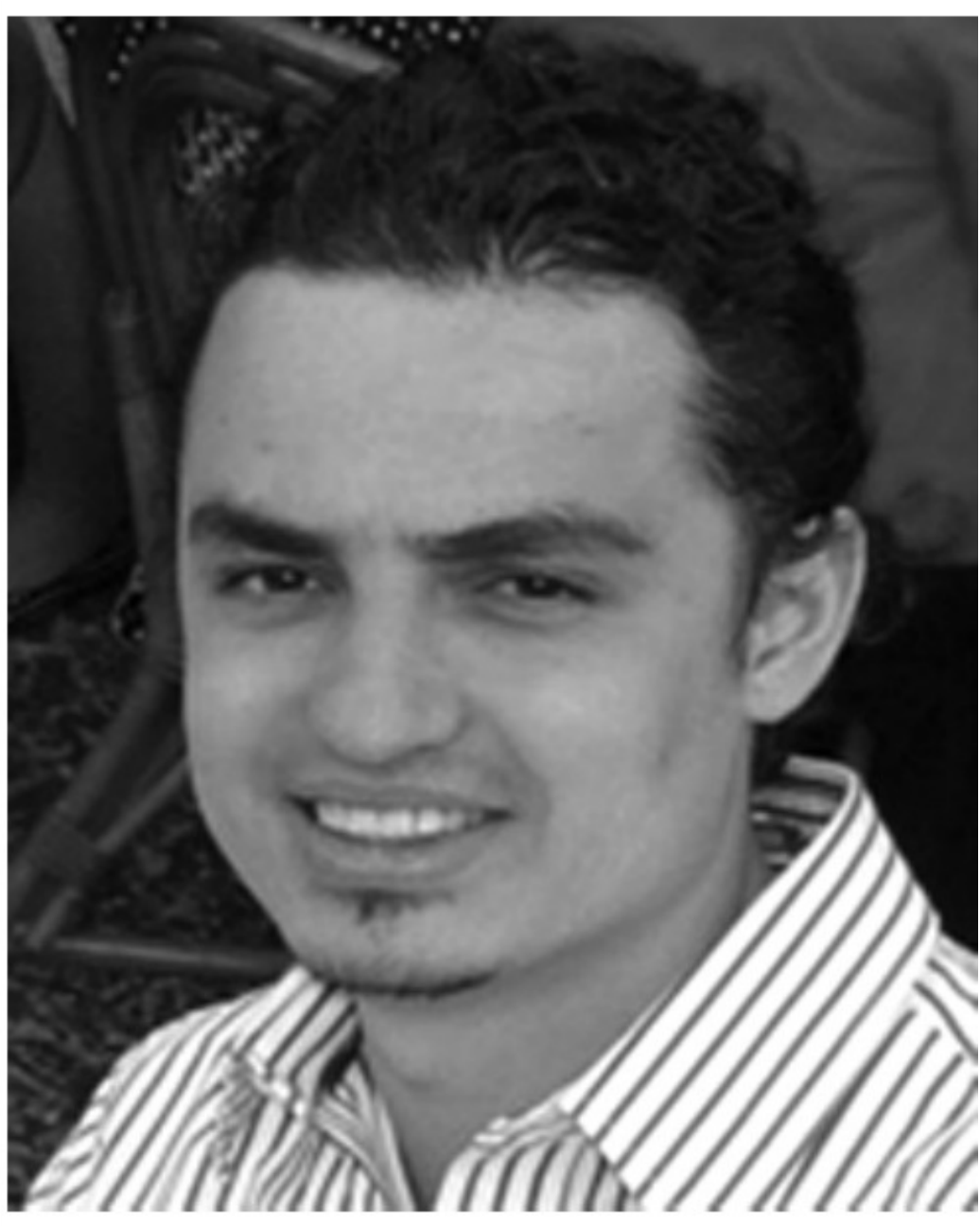}}]{Kemal Ugur}
received the M.S. degree in electrical and computer engineering from the University of British Columbia, Vancouver, BC, Canada, in 2003, and the Ph.D. degree from the Tampere University of Technology, Tampere, Finland, in 2010. He was with Nokia Research Center, Tampere in 2004, where, currently, he is a Principal Researcher and leading a project on next generation video coding technologies. Since he joined Nokia, he has been actively participating in several standardization forums, such as Joint Video Team for the standardization of multiview video coding extension of H.264/AVC, Video Coding Experts Group for exploration toward next generation video coding standard, Joint Collaborative Team on Video Coding for standardization of high efﬁciency video coding standard.
\end{IEEEbiography}
\vspace{-5mm}
\begin{IEEEbiography}[{\includegraphics[width=1in,height=1.25in,clip,keepaspectratio]{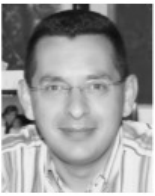}}]{A. Aydin Alatan} received the B.S. degree from Middle East Technical University, Ankara, Turkey, in 1990, the M.S. and DIC degrees from Imperial College of Science, Medicine and Technology, London, U.K., in 1992, and the Ph.D. degree from
Bilkent University, Ankara, Turkey, in 1997, all in electrical engineering.He was a post-doctoral research associate at Center for Image Processing Research at Rensselaer Polytechnic Institute, Troy, NY, between 1997 and 1998 and at the New Jersey Center for Multimedia Research at New Jersey Institute of Technology, Newark, between 1998 and 2000. In August 2000, he joined faculty of Electrical and Electronics Engineering Department at Middle East Technical University.
\end{IEEEbiography}

\end{document}